\documentclass[10pt,twoside,journal]{IEEEtran}
\usepackage{amsfonts,amsmath,amssymb,amsthm}
\usepackage{graphicx,subcaption}
\usepackage{float} 
\usepackage{xspace}
\usepackage{url}
\def\R{{\mathbb{R}}}
\def\RR{{\mathbb{R}^N}}
\def\Z{{\mathbb{Z}}}

\def\balpha{{\boldsymbol{\alpha}}}

\def\bc{{\boldsymbol{c}}}

\def\bg{{\boldsymbol{g}}}

\def\br{{\boldsymbol{r}}}
\def\bu{{\boldsymbol{u}}}
\def\bv{{\boldsymbol{v}}}
\def\bx{{\boldsymbol{x}}}
\def\by{{\boldsymbol{y}}}
\def\bz{{\boldsymbol{z}}}
\def\bn{{\boldsymbol{n}}}
\def\zv{{\boldsymbol{0}}}

\def\bB{{\boldsymbol{B}}}

\def\bD{{\boldsymbol{D}}}

\def\bH{{\boldsymbol{H}}}
\def\bI{{\boldsymbol{I}}}
\def\bL{{\boldsymbol{L}}}
\def\bM{{\boldsymbol{M}}}

\def\bPsi{{\boldsymbol{\Psi}}}

\def\conv{\Gamma_0(\H)}

\def\H{{\mathcal{H}}}

\def\cS{{\mathcal{S}}}
\def\half{{\frac{1}{2}}}

\newcommand{\defeq}{:=}
\newcommand{\bA}{\boldsymbol{A}}

\newcommand{\Frac}[2]{{#1}/{#2} }
\newcommand{\ip}[2]{\left\langle {#1}\, , \,{#2}\right\rangle}
\newcommand{\ipn}[2]{\left\langle {#1} , {#2}\right\rangle}
\newcommand{\set}[2]{\left\{{#1}\ |\ {#2}\right\}}
\newcommand{\norm}[1]{\left\|#1\right\|}
\newcommand{\deri}[2]{\frac{\partial {#1}}{\partial {#2}}}

\makeatletter
\DeclareRobustCommand\onedot{\futurelet\@let@token\@onedot}
\def\@onedot{\ifx\@let@token.\else.\null\fi\xspace}
 
\def\ie{\emph{i.e}\onedot} 
 
\def\etc{\emph{etc}\onedot}

\makeatother
\newtheorem{thm}{Theorem}
\newtheorem{prop}{Proposition}

\DeclareMathOperator{\prox}{prox}
\DeclareMathOperator{\dom}{dom}
\DeclareMathOperator{\gra}{gra}
\DeclareMathOperator{\proj}{proj}
\DeclareMathOperator{\id}{Id}
\DeclareMathOperator{\ran}{ran}
\DeclareMathOperator{\gr}{gra}
\DeclareMathOperator{\T}{T}
\DeclareMathOperator{\diag}{diag}
\DeclareMathOperator*{\argmin}{argmin}
\DeclareMathOperator*{\argmax}{argmax}

\floatstyle{ruled}
\newfloat{algorithm}{t}{loa}
\floatname{algorithm}{Algorithm}

\begin{document}

\title{Learning Convex Regularizers for \\ Optimal Bayesian Denoising%
  \thanks{This work was funded by the Swiss National Science Foundation under
  Grant 200020-162343\@.} \thanks{The authors are with Biomedical Imaging
    Group, \'{E}cole polytechnique f\'{e}d\'{e}rale de Lausanne (EPFL), Station
  17, CH-1015, Lausanne, Switzerland\@.} } 
  
\author{Ha Q.\ Nguyen,  Emrah Bostan, \IEEEmembership{Member, IEEE}, and
Michael Unser, \IEEEmembership{Fellow, IEEE} }

\maketitle
\begin{abstract}
We propose a data-driven algorithm for the maximum a posteriori (MAP)
estimation of stochastic processes from noisy observations. The primary
statistical properties of the sought signal is specified by the penalty
function (i.e., negative logarithm of the prior probability density function).
Our alternating direction method of multipliers (ADMM)-based approach
translates the estimation task into successive applications of the proximal
mapping of the penalty function. Capitalizing on this direct link, we define
the proximal operator as a parametric spline curve and optimize the spline
coefficients by minimizing the average reconstruction error for a given
training set. The key aspects of our learning method are that the associated
penalty function is constrained to be convex and the convergence of the ADMM
iterations is proven.  As a result of these theoretical guarantees, adaptation
of the proposed framework to different levels of measurement noise is extremely
simple and does not require any retraining. We apply our method to estimation
of both sparse and non-sparse models of L\'{e}vy processes for which the
minimum mean square error (MMSE) estimators are available. We carry out a
single training session and perform comparisons at various signal-to-noise
ratio (SNR) values.  Simulations illustrate that the performance of our
algorithm is practically identical to the one of the MMSE estimator
irrespective of the noise power.  
\end{abstract}


\section{Introduction}\label{sec:introduction}
\IEEEPARstart{S}{tatistical} inference is a central theme to the theory of
inverse problems and, in particular, Bayesian methods have been successfully
used in several signal processing
problems~\cite{Tarantola.2005,Ruanaidh.Fitzgerald.2012,Candy.2016}. Among these
is the estimation of signals under the additive white Gaussian noise (AWGN)
hypothesis, which we shall consider throughout this paper. Conventionally, the
unobservable signal is modeled as a random object with a prior probability
density function (pdf) and the estimation is performed by assessing the
posterior pdf that characterizes the problem statistically.  In addition to
being important on its own right, this classical problem has recently gained a
significant amount of interest. The main reason of the momentum is that
Bayesian estimators can be directly integrated---as ``denoisers''---into
algorithms that are designed for more sophisticated inverse
problems~\cite{Venkatakrishnan.etal2013}.  In plain terms, employing a more
accurate denoising technique helps one improve the performance of the
subsequent reconstruction method. Such ideas have been presented in various
applications including deconvolution~\cite{Rond.etal.2016}, super-resolved
sensing~\cite{Chan.etal2017}, and compressive imaging~\cite{Sreehari.etal2016},
to name just a few.  

The MAP inference is by far the most widely used Bayesian paradigm due its
computational convenience~\cite{Kay.1993}.  Compatibility of MAP with the
regularized least-squares approach is well-understood. For example, by this
parallelism, the implicit statistical links between the popular sparsity-based
methods~\cite{RudinOF:1992} and MAP considerations based on generalized
Gaussian, Laplace, or hyper-Laplace priors is
established~\cite{Bouman.Sauer1993,Krishnan.Fergus2009,Babacan.etal2010}.
Moreover, recent iterative optimization techniques including (fast) iterative
shrinkage/thresholding algorithm
((F)ISTA)~\cite{FigueiredoN:2003,BectBAC:2003,DaubechiesDM:2004} and
ADMM~\cite{BoydPCPE:2011} allow us to handle these type of problems very
efficiently.

Fundamentally, the estimation performance of MAP is differentiated by the preferred
prior model. When the inherent nature of the underlying signal is (fully or
partially) deterministic, identification of the right prior is challenging.
Fitting statistical models to such signals (or collections of
them) is feasible~\cite{Choi.Baraniuk1999}. Yet, the apparent downside is that
the reference pdf, which specifies the inference, can be arbitrary.
Even when the signal of interest is purely stochastic and the prior is exactly
known, deviations from the initial statistical assumptions is
observed~\cite{Nikolova.2007}. More importantly, mathematical characterization
of the MAP estimate (i.e., the maximizer of the posterior pdf) by means of
mean-square error (MSE) is available only in limited cases\cite{Stein:1981}.
Hence, algorithms driven by rigorous MAP considerations can still be suboptimal
with respect to MSE~\cite{Gribonval.etal2012,Unser.Tafti2011}. These
observations necessitate revisiting MAP-like formulations from the perspective
of estimation accuracy instead of strict derivations based on the prior model. 

\subsection{Overview of Related Literature} 

Several works have aimed at improving the performance of MAP. Cho~\textit{et
al.} have introduced a nonconvex method to enforce the strict fit between the
signal (or its attributes) and their choice of prior distribution.  Gribonval
has shown that the MMSE can actually be stated as a variational problem that it
is in spirit of MAP~\cite{Gribonval:2011}.  Based on the theory of
continuous-domain sparse stochastic processes~\cite{Unser.Tafti2014},
Amini~\textit{et al.} have analyzed the conditions under which the performance
of MAP can be MSE-optimal~\cite{AminiKBU:2013}.  In~\cite{BostanKNU:2013},
Bostan~\textit{et al.} have investigated the algorithmic implications of
various prior models for the proximal (or the shrinkage) operator that takes
part in the ADMM steps.  Accordingly, Kazerouni~\textit{et al.} and
Tohidi~\textit{et al.} have demonstrated that MMSE performance can be achieved
for certain type of signals if the said proximal operator is replaced with
carefully chosen MMSE-type shrinkage functions~\cite{KazerouniKBU:2013}. Such
methods, however, rely on the full knowledge of the prior model, which
significantly limits their applicability. 

Modification of the proximal operators have also been investigated based on
deterministic principles. In particular, motivated by the outstanding success
of convolutional neural networks (CNNs)~\cite{Lecun.etal2015}, several
researchers have used learning-based methods to identify model parameters (thus
the proximal). In this regard, Gregor and LeCun~\cite{GregorL:2010}, and
Kamilov and Mansour~\cite{KamilovM:2016} have considered sparse encoding
applications and replaced the soft-thresholding step in (F)ISTA with a learned
proximal.  Yang~\textit{et al.} have applied learning to the proximal operator
of ADMM for improved magnetic resonance (MR) image
reconstruction~\cite{YangSHX:2016}.
In~\cite{SchmidtR:2014,ChenYP:2015,ChenP:2015}, learning different shrinkage
functions for each iteration is proposed. A variant of these methods
is considered by Lefkimmiatis~\cite{Lefkimmiatis:2016}. More relevant to the
present context, Samuel and Tappen have learned the model parameters of MAP
estimators for continuous-valued Markov random fields
(MRFs)~\cite{Samuel.Tappen.2009}. What is common in all these techniques is
that the proximal algorithm at hand is trained, which is a nonconvex
optimization problem, without any restrictions.


%
%
%

\subsection{Contributions}
We revisit the MAP problem that is cast as the minimization of a quadratic
fidelity term regularized by a penalty function. The latter captures the statistics of
the original signal. The problem is solved via ADMM by iteratively applying the
proximal operator associated with the penalty function. This direct link
provides us with the proper framework to formulate and rigorously analyze our method.
Our main contributions are summarized as follows: 
\begin{itemize}
  \item Proposal of a new estimator by learning an iteration-independent
    penalty function that is convex. The convexity constraint is appropriately
    characterized in terms of the spline coefficients that parameterize the
    corresponding proximal operator. The learning process optimizes the
    coefficients so that the mean $\ell_2$-normed error between a set of
    ground-truth signals and the ADMM reconstructions (from their noise-added
    versions) is minimized. 

  \item Convergence proof of the resultant ADMM scheme based on the
    above-mentioned convexity confinement. Consequently, the learned penalty
    function is adjusted from one level of noise to another by a simple scaling
    operation, eliminating the need for retraining. Furthermore, assuming
    symmetrically distributed signals, the number of learning parameters is
    reduced by a half. 

  \item Application of the proposed learning framework on two model signals,
    namely the Brownian motion and compound Poisson process. The main reason for choosing these models is that 
    their (optimal) MMSE estimations are available for comparison.
    Furthermore, since these stochastic processes can be decorrelated by the finite difference operator, dictionary learning is no longer needed and we can focus only on the nonlinearity learning. Experiments show
    that, for a wide range of noise variances, ADMM reconstructions with
    learned penalty functions are almost identical to the minimum mean square
    error (MMSE) estimators of these signals. We further demonstrate the
    practical advantages of the proposed learning scheme over its unconstrained
    counterpart. 
\end{itemize}

\subsection{Outline}
In the sequel, we provide an overview of the necessary mathematical tools in
Section~\ref{sec:background}. In Section~\ref{sec:unconstrained}, we present
our spline-based parametrization for the proximal operator and formulate the
unconstrained version of our algorithm. This is then followed by the
introduction of the constraint formulation in terms of the spline coefficients
in Section~\ref{sec:constraint}. We prove the convergence and the scalability
(with respect to noise power) in Section~\ref{sec:constrain}.  Finally,
numerical results are illustrated in Section~\ref{sec:experiments} where we
show that our algorithm achieves the MMSE performance for L\'evy processes with
different sparsity characteristics. 

\section{Background}\label{sec:background}
\subsection{Monotone operator theory}
We review here some notation and background from convex analysis and monotone operator theory; see~\cite{BauschkeC:2011} for further details. Let us restrict ourselves to the Hilbert space $\H = \R^d$, for some dimension $d\geq 1$, equipped with the Euclidean scalar product $\ip{\cdot}{\cdot}$ and norm $\|\cdot\|_2$. The identity operator on $\H$ is denoted by $\id$. Consider a set-valued operator $T:\H \rightarrow 2^{\H}$ that maps each vector $\bx\in\H$ to a set $T\bx\subset \H$. The domain, range, and graph of operator $T$ are respectively defined by
\begin{align*}
\dom T &= \left\{\bx\in\H\ |\ T\bx\neq \emptyset\right\},\\
\ran T &= \left\{\bu\in\H\ |\ (\exists\, \bx\in\H)\, \bu\in T\bx\right\},\\
\gra T &= \left\{(\bx,\bu)\in\H\times \H\ |\ \bu\in T\bx\right\}.
\end{align*}
We say that $T$ is single-valued if $T\bx$ has a unique element for all $\bx\in \dom T$.  
The inverse $T^{-1}$ of $T$ is also a set-valued operator from $\H$ to ${2^{\H}}$ defined by
	\begin{align*}
	T^{-1}\bu\defeq \set{\bx\in\H}{\bu\in T\bx}.
	\end{align*}
It is straightforward that to see that $\dom T =\ran T^{-1}$ and $\ran T =\dom T^{-1}$. $T$ is called \emph{monotone} if
\begin{align*}
\ip{\bx-\by}{\bu-\bv}\geq 0, \quad\forall (\bx,\bu)\in \gra T, \forall (\by,\bv)\in \gra T.
\end{align*}
In 1-D, a monotone operator is simply a non-decreasing function.
$T$ is \emph{maximally monotone} if it is monotone and there exists no monotone operator $S$ such that $\gra T\subsetneqq\gra S$. A handy characterization of the maximal monotonicity is given by Minty's theorem~\cite[Theorem 21.1]{BauschkeC:2011}.
\begin{thm}[Minty]\label{thm:Minty}
A monotone operator $T:\H\rightarrow 2^{\H}$ is maximally monotone if and only if $\ran (\id + T) = \H$.
\end{thm}
For an integer $n\geq 2$, $T:\H\rightarrow 2^{\H}$ is $n$-\emph{cyclically monotone} if, for every $n$ points $(\bx_i,\bu_i)\in \gra T, i=1,\ldots,n$, and for $x_{n+1}=x_1$, we have that 
\begin{align*}
	\sum_{i=1}^{n} \ip{\bx_{i+1}-\bx_{i}}{\bu_i} \leq 0.
\end{align*}
An operator is cyclically monotone if it is $n$-cyclically monotone for all $n\geq 2$. This is a stronger notion of monotonicity because being monotone is equivalent to being $2$-cyclically monotone. Moreover, $T$ is \emph{maximally cyclically monotone} if it is cyclically monotone and there exists no cyclically monotone operator $S$ such that $\gra T\subsetneqq \gra S$. An operator $T$ is said to be \emph{firmly nonexpansive} if
	\begin{align*}
		\langle\bx-\by,\bu-\bv\rangle \geq \|\bu-\bv\|^2 , \,\forall (\bx,\bu)\in \gra T, \forall (\by,\bv)\in \gra T.
	\end{align*}
It is not difficult to see that a firmly nonexpansive operator must be both single-valued and monotone.

We denote by $\Gamma_0(\H)$ the class of all proper lower-semicontinuous \emph{convex} functions $f:\H \rightarrow (-\infty,+\infty]$. For any proper function $f:\H \rightarrow(-\infty,+\infty]$, the \emph{subdifferential} operator $\partial f:\H\rightarrow 2^{\H}$ is defined by
\begin{align*}
\partial f(\bx) = \set{\bu\in\H}{ \langle \by-\bx,\bu \rangle\leq f(\by)-f(\bx) ,\forall \by \in \H},
\end{align*}
whereas, the \emph{proximal} operator $\prox_f:\H\rightarrow 2^{\H}$ is given by 
\begin{align*}
\prox_f(\bx) = \argmin_{\bu\in\H} \left\{f(\bu) + \frac{1}{2}\|\bu-\bx\|_2^2\right\}.
\end{align*}
It is remarkable that, when $f\in\conv$, $\partial f$ is maximally cyclically monotone, $\prox_f$ is firmly nonexpansive, and the two operators are related by
\begin{align}
	\prox_f = \left(\id + \partial f\right)^{-1},
\end{align}
where the right-hand side is also referred to as the \emph{resolvent} of $\partial f$.
Interestingly, any maximally cyclically monotone operator is the subdifferential of some convex function, according to  Rockafellar's theorem~\cite[Theorem 22.14]{BauschkeC:2011}. 
\begin{thm}[Rockafellar]\label{thm:Rocka}
	$A:\H\rightarrow 2^{\H}$ is maximally cyclically monotone if and only if there exists $f\in\conv$ such that $A=\partial f$.
\end{thm}

\subsection{Denoising problem and ADMM}
Let us consider throughout this paper the denoising problem in which a signal $\bx\in\R^N$ is estimated from its corrupted version $\by=\bx+\bn$, where $\bn$ is assumed to be additive white Gaussian noise (AWGN) of  variance $\sigma^2$. An estimator of $\bx$ from $\by$ is denoted by $\hat{\bx}(\by)$. We treat $\bx$ as a random vector generated from the joint probability density function (pdf) $p_{X}$. It is assumed that $\bx$ is \emph{whitenable} by a  matrix $\bL\in\R^{N\times N}$ such that the transformed vector $\bu=\bL\bx$ has identically independently distributed (i.i.d.) entries. The joint pdf $p_U$ of the so-called \emph{innovation} $\bu$ is therefore separable, \emph{i.e.},
\begin{align*}
p_{U}(\bu) = \prod_{i=1}^{N}p_U(u_i),
\end{align*}
where, for convenience, $p_U$ is reused to denote the 1-D pdf of each component of $\bu$. 
We define $\Phi_U(\bu) = -\log p_U (\bu)$ as the \emph{penalty function} of $\bu$. This function is then separable in the sense that 
\begin{align*}
\Phi_{U}(\bu) = \sum_{i=1}^{N}\Phi_U(u_i),
\end{align*}
where $\Phi_U$ is again used to denote the 1-D penalty function of each entry $u_i$.

The MMSE estimator, which is optimal if the ultimate goal is to minimize the expected squared error between the estimate $\hat{\bx}$ and the original signal $\bx$, is given by Stein's formula~\cite{Stein:1981}
\begin{align}\label{eq:Stein}
\hat{\bx}_{\rm MMSE}(\by) = \by + \sigma^2 \nabla\log p_Y(\by), 
\end{align}
where $p_Y$ is the joint pdf of the measurement $\by$ and $\nabla$ denotes the gradient operator. Despite its elegant expression, the MMSE estimator, in most cases, is computationally intractable since $p_Y$ is obtained through a highly dimensional convolution between the prior distribution $p_X$ and the Gaussian distribution $g_{\sigma}(\bn)= (2\pi\sigma^2)^{-N/2}\exp(-\Frac{\norm{\bn}^2}{2\sigma^2})$. However, for L\'{e}vy processes, which have independent and stationary increments, the MMSE estimator is computable using a message passing algorithm~\cite{KamilovPAU:2013}. 

The maximum a posteriori estimator (MAP) is given by
\begin{align}
	\hat{\bx}_{\rm MAP}(\by) &= \argmax_{\bx} p_{X|Y}(\bx|\by)\nonumber\\
							&=  \argmax_{\bx}\left\{ p_{Y|X}(\by|\bx)\,p_{X}(\bx)\right\}\nonumber\\
							&=\argmin_{\bx} \left\{\half\|\by-\bx\|^2_2 + \sigma^2 \Phi_X(\bx)\right\} .\label{eq:MAP1}
\end{align}
where $\Phi_X(\bx) = -\log p_X(\bx)$ is the (nonseparable) penalty function of $\bx$. In other words, the MAP estimator is exactly the proximal operator of $\sigma^2\Phi_X$. Since $\bu=\bL\bx$ is associated with the separable penalty function $\Phi_{U}$, the minimization in~\eqref{eq:MAP1} can be written as
\begin{align}
\min_{\bx} \left\{\half\|\by-\bx\|^2_2 + \sigma^2 \sum_{i=1}^N\Phi_U([\bL\bx]_i)\right\}.\label{eq:MAP2}
\end{align}     
This expression of the MAP reconstruction resembles the conventional regularization-based approach in which the transform $\bL$ is designed to sparsify the signal, the penalty function $\Phi_U$ is chosen---the typical choice being the $\ell_1$-norm---to promote the sparsity of the transform coefficients $\bu$. The parameter $\sigma^2$ is set (not necessarily to the noise variance) to trade off the quadratic fidelity term with the regularization term. The optimization problem~\eqref{eq:MAP2} can be solved efficiently by iterative algorithms such as the alternating direction method of multipliers (ADMM). To that end, we form the augmented Langrangian 
\begin{align*}
	\half\|\by-\bx\|^2_2 + 	\sigma^2\Phi_U(\bu) - \langle\balpha,\bL\bx-\bu\rangle + \frac{\mu}{2} \|\bL\bx-\bu\|^2_2					
\end{align*}
and successively mimimize this functional with respect to each of the variables $\bx$ and $\bu$, while fixing the other one; the Lagrange multiplier $\balpha$ is also updated appropriately at each step. In particular, at iteration $k+1$, the updates look like
\begin{align}
\bx^{(k+1)} &= \left(\bI+\mu\bL^{\T}\bL\right)^{-1} \left(\by+\bL^{\T}\left(\mu\bu^{(k)}+\balpha^{(k)}\right)\right)\label{eq:step1}\\
\balpha^{(k+1)} &= \balpha^{(k)} - \mu\left(\bL\bx^{(k+1)}-\bu^{(k)}\right)\label{eq:step2}\\
\bu^{(k+1)} &= \prox_{\sigma^2/\mu\Phi_U}\left(\bL\bx^{(k+1)}-\frac{1}{\mu}\balpha^{(k+1)}\right)\label{eq:step3}.
\end{align}
Here, $\bu$ and $\balpha$ are initialized to be $\bu^{(0)}$ and $\balpha^{(0)}$, respectively; $\bI\in \R^{N\times N}$ denotes the identity matrix.
If the proximal operator $\prox_{\sigma^2/\mu\Phi_U}$ in~\eqref{eq:step3} is replaced with a general operator $T$, we refer to the above algorithm as the \emph{generalized ADMM} associated with $T$. When the operator $T$ is separable, \ie, $T(\bu) = \left(T(u_1),\ldots,T(u_N)\right)$, we refer to the 1-D function $T:\R\rightarrow \R$ as the \emph{shrinkage function}; the name comes from the observation that typical proximal operators, such as the soft-thresholding, shrink large values of the input in a pointwise manner to reduce the noise. In what follows, we propose a learning approach to the denoising problem in which the shrinkage function $T$ of the generalized ADMM is optimized in the MMSE sense from data, instead of being engineered as in sparsity-promoting schemes.


\section{Learning Unconstrained Shrinkage Functions}\label{sec:unconstrained}
\subsection{Learning algorithm}
To learn the shrinkage function $T:\R\rightarrow \R$, we parameterize it via a  spline representation:
\begin{align}\label{eq:spline}
T(x)= \sum_{m=-M}^{M}c_m\psi\left(\frac{x}{\Delta}-m\right),
\end{align}
where $\psi$ is some kernel (radial basis functions, B-splines, \etc) and $\Delta$ is the sampling step size that defines the distant between consecutive spline knots. We call such function $T$ a \emph{shrinkage spline}.
Consider the generalized ADMM associated with $T$. Fix the transform matrix $\bL$, the penalty parameter $\mu$, and the number of ADMM iterations $K$. The  vector $\bc\in\R^{2M+1}$ of spline coefficients of $T$ is to be learned by minimizing the following cost function:
\begin{align}\label{eq:cost}
J(\bc) = \half\sum_{\ell=1}^{L}\norm{\bx^{(K)}(\bc,\by_\ell)-\bx_{\ell}}^2_2, 
\end{align}
where $\{\bx_{\ell}\}_{\ell=1}^L$ is the collection of $L$ ground-truth signals from which the observations $\{\by_{\ell}\}_{\ell=1}^L$ are made and $\left\{\bx^{(K)}(\by_\ell)\right\}_{\ell=1}^L$ are the corresponding reconstructions from $\{\by_{\ell}\}_{\ell=1}^L$ using the generalized ADMM with $K$ iterations. For notational simplicity, from now on we drop the subscript $\ell$ and develop a learning algorithm for a single training example $(\bx,\by)$ that can be easily generalized to  training sets of arbitrary size. The cost function is thus simplified to
\begin{align}\label{eq:cost_func}
J(\bc) = \half\norm{\bx^{(K)}(\bc,\by)-\bx}^2_2. 
\end{align}
The minimization of this function is performed via a simple gradient descent that is described in Algorithm~\ref{alg:learn_uncons}; we call the algorithm unconstrained learning to distinguish it from the constrained learning that will be presented later. As in every neural network, the computation of the gradient of the cost function is performed in a backpropagation manner, which will be detailed in the next section.
\begin{algorithm}
	\caption{Unconstrained Learning}
	\label{alg:learn_uncons}
	\vspace*{3mm}
	{\bf Input:} training example $(\bx,\by)$, learning rate $\gamma>0$, sampling step $\Delta$, number of spline knots $2M+1$.\\
	{\bf Output:} spline coefficients  $\bc^{*}$. 
\begin{enumerate}
	\item \emph{Initialize}: Set $0\leftarrow i$, choose $\bc^{(0)}\in\R^{2M+1}$.
	\item Compute the gradient $\nabla J\left(\bc^{(i)}\right)$ via Algorithm~\ref{alg:grad_comput}.
	\item Update $\bc$ as: 
	$$
		\bc^{(i+1)} = \bc^{(i)} - \gamma \nabla J\left(\bc^{(i)}\right).
	$$
	\item Return $\bc^* = \bc^{(i+1)}$ if a stopping criterion is met, otherwise set $i\leftarrow i+1$ and go to step 2.
\end{enumerate}
\end{algorithm}

\subsection{Gradient computation}
We devise in this section a backpropagation algorithm to evaluate the gradient of the cost function with respect to the spline coefficients of the shrinkage function. We adopt the following convention for matrix calculus: for a function $y:\R^{m}\rightarrow\R$ of vector variable $\bx$, its gradient is a column vector given by
\begin{align*}
\nabla y(\bx)=\deri{y}{\bx} = 
\begin{bmatrix}
\deri{y}{x_1} &\deri{y}{x_2} &\cdots& \deri{y}{x_m}
\end{bmatrix}
^{\T},
\end{align*}
whereas, for a vector-valued function $\by:\R^{m}\rightarrow\R^{n}$ of vector variable $\bx$, its Jacobian is an $m\times n$ matrix defined by
\begin{align*}
\deri{\by}{\bx} = 
\begin{bmatrix}
	\deri{y_1}{\bx} & \deri{y_2}{\bx} & \cdots & \deri{y_n}{\bx} 
\end{bmatrix} 
=
\begin{bmatrix}
\deri{y_1}{x_1} &  \cdots & \deri{y_n}{x_1}\\
\vdots &  \ddots & \vdots\\
\deri{y_1}{x_m} &  \cdots & \deri{y_n}{x_m}
\end{bmatrix}.
\end{align*}

We are now ready to compute the gradient of the cost function $J$ with respect to the parameter vector $\bc$. For simplicity, for $k=0,\ldots,K-1$, put
\begin{align*}
\bM &= (\bI +\mu\bL^{\T}\bL)^{-1},\\
\bz^{(k+1)}&=\by+\bL^{\T}\left(\mu\bu^{(k)}+\balpha^{(k)}\right),\\
\bv^{(k+1)}&= \bL\bx^{(k+1)}-\frac{1}{\mu}\balpha^{(k+1)}.
\end{align*}
By using these notations, we concisely write the updates at iteration $k+1$ of the generalized ADMM associated with operator $T$ as
\begin{align*}
	\bx^{(k+1)} &= \bM \bz^{(k+1)},\\
	\balpha^{(k+1)} &= \balpha^{(k)} - \mu\left(\bL\bx^{(k+1)}-\bu^{(k)}\right),\\
	\bu^{(k+1)} &= T\left(\bv^{(k+1)}\right).
\end{align*}
First, applying the chain rule to~\eqref{eq:cost_func} yields
\begin{align}\label{eq:grad_first}
 \nabla J(\bc)  &= \frac{\partial \bx^{(K)}}{\partial \bc}\frac{\partial J}{\partial \bx^{(K)}}
 = \frac{\partial \bx^{(K)}}{\partial \bc} \left(\bx^{(K)}-\bx\right)
\end{align}
Next, from the updates of the ADMM and by noting that $\bL$ and $\by$ does not depend on $\bc$, for $k=0,\ldots,K-1$, we get
\begin{align*}
\frac{\partial \bx^{(k+1)}}{\partial \bc}&= \deri{\bz^{(k+1)}}{\bc}\bM^{\T}= \left(\mu\deri{\bu^{(k)}}{\bc} + \deri{\balpha^{(k)}}{\bc} \right)\bL\bM,
\end{align*}
\begin{align*}
\frac{\partial \balpha^{(k)}}{\partial \bc}&=\frac{\partial \balpha^{(k-1)}}{\partial \bc} -\mu \deri{\bx^{(k)}}{\bc}\bL^{\T} + \mu \deri{\bu^{(k-1)}}{\bc},
\end{align*}
and
\begin{align*}
\frac{\partial \bu^{(k)}}{\partial \bc} &= \deri{\bv^{(k)}}{\bc}\deri{\bu^{(k)}}{\bv^{(k)}} + \deri{\bc}{\bc}\deri{\bu^{(k)}}{\bc} \\
&= \left(\deri{\bx^{(k)}}{\bc}\bL^{\T}-\frac{1}{\mu}\deri{\balpha^{(k)}}{\bc}\right)\bD^{(k)} + \bPsi^{(k)},
\end{align*}
where $\bD^{(k)}=\diag\left(T'(\bv^{(k)})\right)$ is the diagonal matrix whose entries on the diagonal are the derivatives of $T$ at $\{v^{(k)}_i\}_{i=1}^N$, and $\bPsi^{(k)}$ is a matrix defined by
\begin{align*}
\Psi^{(k)}_{ij} &= \psi\left(\frac{v^{(k)}_j}{\Delta} - i\right).
\end{align*}
Proceeding with simple algebraic manipulation, we arrive at
\begin{subequations}\label{eq:recursion}
	\begin{align}
		\deri{\bx^{(k+1)}}{\bc} = \left(\deri{\balpha^{(k)}}{\bc} +\mu \deri{\bu^{(k)}}{\bc}\right)\bA,
	\end{align}
	\begin{align}
		\deri{\balpha^{(k)}}{\bc} +\mu \deri{\bu^{(k)}}{\bc} =  \left(\deri{\balpha^{(k-1)}}{\bc} +\mu \deri{\bu^{(k-1)}}{\bc}\right)\bB^{(k)} + \mu\bPsi^{(k)}.
	\end{align}
\end{subequations}
where
\begin{align*}
\bA &= \bL\left(\bI +\mu\bL^{\T}\bL\right)^{-1},\\
\bB^{(k)} &= \bI - \mu \bA\bL^T + \left(2\mu\bA\bL^{\T}-\bI\right)\bD^{(k)}.
\end{align*}

Finally, by combining~\eqref{eq:grad_first} with~\eqref{eq:recursion} and by noting that $\Frac{\partial\balpha^{(0)}}{\partial\bc}=\Frac{\partial\bu^{(0)}}{\partial\bc}=\zv$, we propose a backpropagation algorithm to compute the gradient of the cost function $J$ with respect to the spline coefficients $\bc$ as described in Algorithm~\ref{alg:grad_comput}. We refer to the generalized ADMM that uses a shrinkage function learned via Algorithm~\ref{alg:learn_uncons} as MMSE-ADMM.

\begin{algorithm}
	\caption{Backpropagation for unconstrained learning}
	\label{alg:grad_comput}
	\vspace*{3mm}
		{\bf Input:} signal $\bx\in\R^N$, measurement $\by\in\R^{N}$, transform matrix $\bL\in\R^{N\times N}$, kernel $\psi$, sampling step $\Delta$, number of spline knots $2M+1$, current spline coefficients $\bc\in\R^{2M+1}$, number of ADMM iterations $K$.\\
		{\bf Output:} gradient  $\nabla J(\bc)$.
\begin{enumerate}
	\item Define:
	\begin{align*}
		\psi_i&=\psi(\cdot/\Delta - i), \text{ for } i=-M,\ldots,M\\
		\bA &= \bL\left(\bI +\mu\bL^{\T}\bL\right)^{-1}
	\end{align*} 
	\item Run $K$ iterations of the generalized ADMM with the shrinkage spline $T=\sum_{i=-M}^{M}c_i\psi_i$. Store $\bx^{(K)}$ and, for all $k=1,\ldots,K$, store
	\begin{align*}
	\bv^{(k)}&=\bL\bx^{(k)}-\balpha^{(k)}/\mu,\\
	\bPsi^{(k)} &= \left\{\psi_i\left(v^{(k)}_j\right)\right\}_{i,j},\\
	\bB^{(k)} &= \bI -\mu\bA\bL^{\T} + \left(2\mu\bA\bL^{\T}-\bI\right)\diag (T' (\bv^{(k)})).
	\end{align*}
	\item \emph{Initialize}: $\br=\bA(\bx^{(K)}-\bx),\bg=\zv, k = K-1$.
	\item Compute: 
	\begin{align*}
	\bg &\leftarrow  \bg +  \mu\bPsi^{(k)}
	\br,\\ 
	\br &\leftarrow \bB^{(k)}\br. 
	\end{align*}
	\item If $k=1$, return $\nabla J(\bc) =\bg$, otherwise, set $k\leftarrow k-1$ and go to step 4.
\end{enumerate}
\end{algorithm}

\section{Learning Constrained Shrinkage Functions}\label{sec:constraint}
We propose two constraints for learning the shrinkage functions: firm nonexpansiveness and antisymmetry. The former is motivated by the well-known fact that the proximal operator of a convex function must be firmly nonexpansive~\cite{BauschkeC:2011}; the latter is justified by Theorem~\ref{thm:symmetry}: symmetrically distributed signals imply antisymmetric proximal operator and vice versa.
\begin{thm}\label{thm:symmetry}
	Let $\Phi\in\Gamma_0(\RR)$. $\Phi$ is symmetric if and only if $\prox_{\Phi}$ is antisymmetric.
\end{thm}
\begin{IEEEproof} 
	See Appendix~\ref{app:symmetry}.
\end{IEEEproof}
In order to incorporate the firmly nonexpansive constraint into the learning of spline coefficients, we choose the kernel $\psi$ in the representation~\eqref{eq:spline} to be a B-spline of some integer order. Recall that the B-spline $\beta^n$ of integer order $n\geq 0$ is defined recursively as
\begin{align*}
\beta^0 (x)&=\begin{cases}
1, & |x|\leq 1/2\\
0, & |x|>1/2,
\end{cases} \\
\beta^{n} &= \beta^{n-1} * \beta^{0}, \quad n\geq 1.
\end{align*}
These are compactly supported kernels with many good properties~\cite{Unser:1999}. More importantly, as pointed out in Theorem~\ref{thm:spline}, by using B-spline kernels the firm nonexpansiveness of a shrinkage spline is satisfied as long as its coefficients obey a simple linear constraint.
\begin{thm}\label{thm:spline}
	Let  $\Delta > 0$ and let $\beta^{n}$ be the B-spline of order $n\geq 1$. If $c$ is a sequence such that $0\leq c_m-c_{m-1}\leq \Delta,\forall m\in\Z$, then $f=\sum_{m\in\Z}c_m\beta^{n}(\cdot/\Delta-m)$  is a firmly nonexpansive function.
\end{thm}
\begin{IEEEproof}
	Since $f$ is a 1-D function, it is easy to see that $f$ is firmly expansive if and only if
	\begin{align}\label{eq:target}
	0\leq f(x)-f(y)\leq x-y,\quad \forall x > y.
	\end{align}
	We now show~\eqref{eq:target} by considering 2 different cases.
	
	\underline{$n=1$:} $\beta^{1}$ is the triangle function, and so $f$ is continuous and piecewise-linear. If  $x,y\in[(m-1)\Delta,m\Delta] $ for some $m\in\Z$, then 
	\begin{align*}
	0\leq\frac{f(x)-f(y)}{x-y}=\frac{c_{m}-c_{m-1}}{\Delta}\leq 1,
	\end{align*}
	which implies
	\begin{align}\label{eq:piece}
	0\leq f(x)-f(y)\leq x-y,\quad \forall (m-1)\Delta\leq y < x \leq m\Delta.
	\end{align}
	Otherwise, there exist $k,\ell\in \Z$ such that $y\in [(k -1)\Delta,k\Delta]$, $x\in [\ell\Delta,(\ell+1)\Delta]$. Then, we write
	\begin{align*}
	f(x)-f(y) &= [f(x) -f(\ell\Delta)] + [f(\ell\Delta)-f((\ell-1)\Delta)]\\
	& +\cdots + [f((\ell+1)\Delta)-f(\ell\Delta)] + [f(k)-f(y)].
	\end{align*}
	
	By applying~\eqref{eq:piece} to each term of the above sum, we obtain the desired pair of inequalities in~\eqref{eq:target}, which implies the firm nonexpansiveness of $f$.
	\\
	\underline{$n\geq 2$:}
	$\beta^n$ is now differentiable and so is $f$. Thus, by using the mean value theorem, \eqref{eq:target} is achieved if the derivative $f'$ of $f$ is bounded between 0 and 1, which will be shown subsequently.
	Recall that the derivative of $\beta^{n}$ is equal to the finite difference of $\beta^{n-1}$. In particular,
	\begin{align}
	(\beta^{n})'(x) = \beta^{n-1}\left(x+\frac{1}{2}\right) - \beta^{n-1}\left(x-\frac{1}{2}\right).
	\end{align}
	Hence, for all $x\in\R$,
	\begin{align*}
	&f'\left(x\right)= \frac{1}{\Delta} \sum_{m\in\Z}c_{m}(\beta^{n})'\left(\frac{x}{\Delta}-m\right)\\ 
	& =\frac{1}{\Delta}\sum_{m\in\Z}c_{m}\left\{\beta^{n-1}\left(\frac{x}{\Delta}-m+\frac{1}{2}\right)\right.\\ 
	& \hspace{3cm} -\left.\beta^{n-1}\left(\frac{x}{\Delta}-m-\frac{1}{2}\right)\right\}\\
	&= \frac{1}{\Delta}\sum_{m\in\Z}c_{m}\beta^{n-1}\left(\frac{x}{\Delta}-m+\frac{1}{2}\right) \\
	&\quad - \frac{1}{\Delta}\sum_{m\in\Z}c_{m-1}\beta^{n-1}\left(\frac{x}{\Delta}-m+\frac{1}{2}\right)\text{ (change of variable)}\\
	& = \frac{1}{\Delta}\sum_{m\in\Z}(c_{m}-c_{m-1})\beta^{n-1}\left(\frac{x}{\Delta}+\frac{1}{2}-m\right).
	\end{align*}
	Since $0\leq c_{m}-c_{m-1}\leq \Delta,\forall m\in\Z$ and since $\beta^{n-1}(\Frac{x}{\Delta}+\Frac{1}{2}-m)\geq 0, \forall x\in\R,m\in\Z$, one has the following pair of inequalities for all $x\in\R$:
	\begin{align}
	0\leq f'(x)\leq  \sum_{m\in\Z}\beta^{n-1}\left(\frac{x}{\Delta}+\frac{1}{2}-m\right).\label{eq:pair}
	\end{align}
	By using the partition-of-unity property of the B-spline $\beta^{n-1}$, \eqref{eq:pair} is simplified to
	\begin{align*}
	0\leq f'(x)\leq 1 ,\quad\forall x\in\R,
	\end{align*}
	which finally proves that $f$ is a firmly nonexpansive function. 
\end{IEEEproof}
With the above results, we easily design an algorithm for learning antisymmetric and firmly nonexpansive shrinkage functions. Algorithm~\ref{alg:learn_cons} is the constrained counterpart of Algorithm~\ref{alg:learn_uncons}: the gradient descent is replaced with a projected gradient descent where, at each update, the spline coefficients are projected onto the linear-constraint set described in Theorem~\ref{thm:spline} (this projection is performed via a quadratic programming). The gradient of the cost function in this case is evaluated through Algorithm~\ref{alg:grad_comput_con}, which is just slightly modified from Algorithm~\ref{alg:grad_comput} to adapt to the antisymmetric nature of the shrinkage splines. We refer to the generalized ADMM that uses a shrinkage function learned by Algorithm~\ref{alg:learn_cons} as MMSE-CADMM, where the letter `C' stands for `convex.' The convexity of this learning scheme will be made clear in Section~\ref{sec:constrain}.
\begin{algorithm}
	\caption{Constrained Learning}
	\label{alg:learn_cons}
	\vspace*{3mm}
	{\bf Input:} training example $(\bx,\by)$, learning rate $\gamma>0$, sampling step $\Delta$, number of spline knots $2M+1$.\\
	{\bf Output:} spline coefficients $\bc^{*}$. 
	\begin{enumerate}
		\item Define the linear constraint set
		$$
			\cS =\left\{\bc\in\R^{M}\,|\,0\leq c_m- c_{m-1}\leq \Delta ,\forall m=2,\ldots,M \right\}
		$$
		\item \emph{Initialize}: Set $0\leftarrow i$, choose $\bc^{(0)}\in\cS$.
		\item Compute the gradient $\nabla J\left(\bc^{(i)}\right)$ via Algorithm~\ref{alg:grad_comput}.
		\item Update $\bc$ as: 
		$$
		\bc^{(i+1)} = \proj_{\cS}\left(\bc^{(i)} - \gamma \nabla J\left(\bc^{(i)}\right)\right).
		$$
		\item Return $\bc^* = \bc^{(i+1)}$ if a stopping criterion is met, otherwise set $i\leftarrow i+1$ and go to step 3.
	\end{enumerate}
\end{algorithm}

\begin{algorithm}
	\caption{Backpropagation for constrained learning}
	\label{alg:grad_comput_con}
	\vspace*{3mm}
	{\bf Input:} signal $\bx\in\R^N$, measurement $\by\in\R^{N}$, transform matrix $\bL\in\R^{N\times N}$, B-spline $\psi=\beta^{n}$, sampling step $\Delta$, number of spline knots $2M+1$, current spline coefficients $\bc\in\R^{M}$, number of ADMM iterations $K$.\\
	{\bf Output:} gradient  $\nabla J(\bc)$.
	\begin{enumerate}
		\item Define:
		 \begin{align*}
		 \tilde{\psi}_i &=\psi(\cdot/\Delta-i) - \psi(\cdot/\Delta+i), \text{ for } i=,\ldots,M\\
		 \bA &= \bL\left(\bI +\mu\bL^{\T}\bL\right)^{-1}.
		 \end{align*} 
		\item Run $K$ iterations of the generalized ADMM with the \emph{antisymmetric} shrinkage spline $T = \sum_{i=1}^M c_i\tilde{\psi}_i$. Store $\bx^{(K)}$ and, for all $k=1,\ldots,K$, store
		\begin{align*}
		\bv^{(k)}&=\bL\bx^{(k)}-\balpha^{(k)}/\mu,\\
		\bPsi^{(k)} &= \left\{\tilde{\psi}_i\left(v^{(k)}_j\right)\right\}_{i,j},\\
		\bB^{(k)} &= \bI -\mu\bA\bL^{\T} + \left(2\mu\bA\bL^{\T}-\bI\right)\diag (T' (\bv^{(k)})).
		\end{align*}
		\item \emph{Initialize}: $\br=
		\bA^{\T}(\bx^{(K)}-\bx),\bg=\zv,k=K-1$.
		\item Compute: 
		\begin{align*}
		\bg &\leftarrow  \bg +  \mu\bPsi^{(k)}
		\br,\\ 
		\br &\leftarrow \bB^{(k)}\br. 
		\end{align*}
		\item If $k=1$, return $\nabla J(\bc) =\bg$, otherwise, set $k\leftarrow k-1$ and go to step 4.
	\end{enumerate}
\end{algorithm}

\section{Advantages of Adding Constraints}\label{sec:constrain}
It is clear that imposing the antisymmetric constraint on the shrinkage function reduces the dimension of the optimization problem by a half and therefore substantially reduces the learning time. In this section, we demonstrate, from the theoretical point of view, the two important advantages of imposing the firmly nonexpansive constraint on the shrinkage function: convergence guarantee and scalability with noise level. Thanks to these properties, our learning-based denoiser behaves like a MAP estimator with some convex penalty function that is now different from the conventional penalty function. On the other hand, as experiments later show, the constrained learning scheme nearly achieves the optimal denoising performance of the MMSE estimator.
\subsection{Convergence guarantee}
The following result asserts that the ADMM denoising converges, no matter what the noisy signal is, if it uses a separable firmly nonexpansive operator in the place of the conventional proximal operator. Interestingly, as will be shown in the proof, any separable firmly nonexpansive operator is the proximal operator of a separable convex penalty function. We want to emphasize that the separability is needed to establish this connection, although the reverse statement is known to hold in the multidimensional case~\cite{BauschkeC:2011}.
\begin{thm}\label{thm:converge}
	If $T:\R\rightarrow\R$ is a 1-D firmly nonexpansive function such that $\dom T = \R$, then, for every input $\by\in\R^N$, the reconstruction sequence $\{\bx^{(k)}\}$ of the generalized ADMM associated with the separable operator $T$ converges to 
	\begin{align*}
	\bx^{*} = \argmin_{\bx\in\RR} \left\{\frac{1}{2}\|\by-\bx\|_2^2 + \sum_{i=1}^{N} \Phi([\bL\bx]_i)\right\},
	\end{align*}
	as $k\rightarrow\infty$, where $\Phi\in\Gamma_0(\R)$ is some 1-D convex function.
\end{thm}
\begin{IEEEproof}
	First, we show that there exists  a function $\Phi\in\Gamma(\R)$ such that $T=\prox_{\Phi}$. To that end, let us define 
	\begin{align*}
	S = (T^{-1} -\id).
	\end{align*} 
	The firm nonexpansiveness of $T$ then implies
	\begin{align*}
	&(x-y)(u-v) \geq (x-y)^2,\qquad\, \forall u\in T^{-1}x,v\in T^{-1}y\\
	\Leftrightarrow &(x-y)((u-x)-(v-y)) \geq 0,\ \forall u\in T^{-1}x,v\in T^{-1}y\\
	\Leftrightarrow &(x-y)(\tilde{u}-\tilde{v}) \geq 0,\qquad\qquad\quad\ \; \forall \tilde{u}\in Sx,\tilde{v}\in Sy,
	\end{align*}
	which means that $S$ is a monotone operator. Furthermore, we have that
	\begin{align*}
	\ran (S+\id) = \ran \left(T^{-1}\right) = \dom T = \R.
	\end{align*}
	Therefore, $S$ is maximally monotone thanks to Minty's theorem (Theorem~\ref{thm:Minty}). Since $S$ is an operator on $\R$, we invoke~\cite[Thm. 22.18]{BauschkeC:2011} to deduce that $S$ must also be maximally cyclically monotone. Now, as a consequence of Rockafellar's theorem (Theorem~\ref{thm:Rocka}), there exists a function $\Phi\in\Gamma(\R)$ such that $\partial\Phi = S$. By the definition of $S$ and of $\prox_\Phi$, we have that 
	\begin{align*}
	T = (\id + S )^{-1} = (\id+\partial\Phi) ^{-1} = \prox_{\Phi}.
	\end{align*}
	
	Next, define the cost function
	\begin{align}\label{eq:underlying_cost}
	f(\bx) = \frac{1}{2}\|\by-\bx\|_2^2 + \sum_{i=1}^{N} \Phi([\bL\bx]_i).
	\end{align}
	Replacing $T$ with $\prox_\Phi$, the generalized ADMM associated with $T$ becomes the regular ADMM associated with the above cost function $f$. By using the convexity of the $\ell_2$-norm and of the function $\Phi$, it is well known~\cite[Section 3.2.1]{BoydPCPE:2011} that $f\left(\bx^{(k)}\right)\rightarrow p^*$ as $k\rightarrow\infty$, where $p^*$ is the minimum value of $f$. 
	
	Finally, notice that the function $f$ defined in~\eqref{eq:underlying_cost} is strongly convex. Thus, there exists a unique minimizer $\bx^{*}\in\R^N$ such that $f(\bx^{*})=p^{*}$. Moreover, $\bx^{*}$ is known to be a strong minimizer~\cite[Lemma 2.26]{PlanidenW:2016} in the sense that the convergence of $\left\{f\left(\bx^{(k)}\right)\right\}_k$ to $f(\bx^*)$ implies the convergence of $\left\{\bx^{(k)}\right\}_k$ to $\bx^*$. This completes the proof. 	
\end{IEEEproof}
\subsection{Scalability with noise level}
In all existing learning schemes, the shrinkage function is learned for a particular level of noise and then applied in the reconstruction of testing signals corrupted with the same level of noise. If the noise variance changes, the shrinkage function must be relearned from scratch, which will create a computational burden on top of the ADMM reconstruction. This drawback is due to the unconstrained learning strategy in which the shrinkage function is not necessarily the proximal operator of any function. Despite its flexibility, arbitrary nonlinearity no longer goes hand-in-hand with a regularization-based minimization (MAP-like estimation). By contrast, our constrained learning scheme maintains a connection with the underlying minimization regularized by a convex penalty function. This strategy allows us to easily adjust the learned shrinkage function from one level of noise to another by simply scaling the corresponding penalty function by the ratio between noise variances like the conventional MAP estimator. Proposition~\ref{thm:scale_convex} provides a useful formula to extrapolate the proximal operator of a scaled convex function from the proximal operator of the original function. 
\begin{prop}\label{thm:scale_convex}
	For all $f\in\Gamma_{0}(\RR)$,
	\begin{align}\label{eq:scale}
	\prox_{\lambda f} = \left(\lambda \prox_f^{-1}+ (1-\lambda)\id\right)^{-1},\quad \forall \lambda \geq 0.
	\end{align}
\end{prop}
\begin{IEEEproof}
Recall a basic result in convex analysis~\cite{Rockaffellar:1997} that $\partial (\lambda f) = \lambda \partial f$, for all $f\in\Gamma_0(\RR)$ and for all $\lambda\geq 0$ . Also recall that the proximal operator of a convex function is the resolvent of the subdifferential operator. Therefore,
\begin{align*}
	\prox_{\lambda f} &= \left(\id + \partial(\lambda f)\right)^{-1}=\left(\id + \lambda\partial f\right)^{-1}\\
					 &= \left(\id + \lambda\left(\prox_f^{-1}-\id\right)\right)^{-1}\\
					 &= \left(\lambda \prox_f^{-1}+ (1-\lambda)\id\right)^{-1},
\end{align*} 
completing the proof. 
\end{IEEEproof}

The next result establishes that all members of the family generated by~\eqref{eq:scale} are firmly nonexpansive when the generator $\prox_f$ is replaced with a general firmly nonexpansive operator. It is noteworthy that the result holds in the multidimensional case where a firmly nonexpansive operator is not necessarily the proximal operator of a convex function.
\begin{thm}\label{thm:scale}
	If $T:\RR\rightarrow \RR$ is a firmly nonexpansive operator such that $\dom T =\RR$, then, for all $\lambda\geq 0$, $T_{\lambda}= (\lambda T^{-1} + (1-\lambda)\id)^{-1}$ is firmly nonexpansive  and $\dom T_{\lambda}=\RR$ as well. 
\end{thm}
\begin{IEEEproof}
	The claim is trivial for $\lambda=0$. Assume from now on that $\lambda>0$. We first show that $\dom T_{\lambda}=\RR$. By a straightforward extension of the argument in the proof of Theorem~\ref{thm:converge} to the multidimensional case, we easily have that the operator $S=T^{-1}-\id$ is maximally monotone. It follows that $\lambda S$ is also maximally monotone for all $\lambda >0$. By applying Minty's theorem to the operator $\lambda S$, we obtain
	\begin{align*}
		\dom T_{\lambda} = \ran (\lambda T^{-1}+(1-\lambda)\id) = \ran (\id + \lambda S) = \RR.
	\end{align*}
	 
	 Next, we show that $T_{\lambda}$ is firmly nonexpansive. Let $\bx,\by\in\RR$ and $\bu\in T_\lambda(\bx),\bv\in T_\lambda(\by)$. By the definition of $T_\lambda$, one readily verifies that 
	\begin{align*}
	\bu&= T\left(\frac{\bx+(\lambda-1)\bu}{\lambda}\right),	\quad \bv= T\left(\frac{\by+(\lambda-1)\bv}{\lambda}\right).
	\end{align*}
	The firm nonexpansiveness of $T$ yields
	\begin{align*}
	\|\bu-\bv\|^2&\leq \ip{\frac{\bx+(\lambda-1)\bu}{\lambda}-\frac{\by+(\lambda-1)\bu}{\lambda}}{\bu-\bv}\\
	&= \frac{\lambda-1}{\lambda}\|\bu-\bv\|^2  + \frac{1}{\lambda}\ip{\bx-\by}{\bu-\bv},
	\end{align*}
	which translates to
	\begin{align*}
	\|\bu-\bv\|^2&\leq \ip{\bx-\by}{\bu-\bv}.
	\end{align*}
	This confirms that $T_{\lambda}$ is a firmly nonexpansive operator. 
\end{IEEEproof}

\section{Experimental Results}\label{sec:experiments}
\begin{figure}
	\centering
	\begin{subfigure}[b]{0.9\linewidth}
		\centering
		\includegraphics[width=\linewidth]{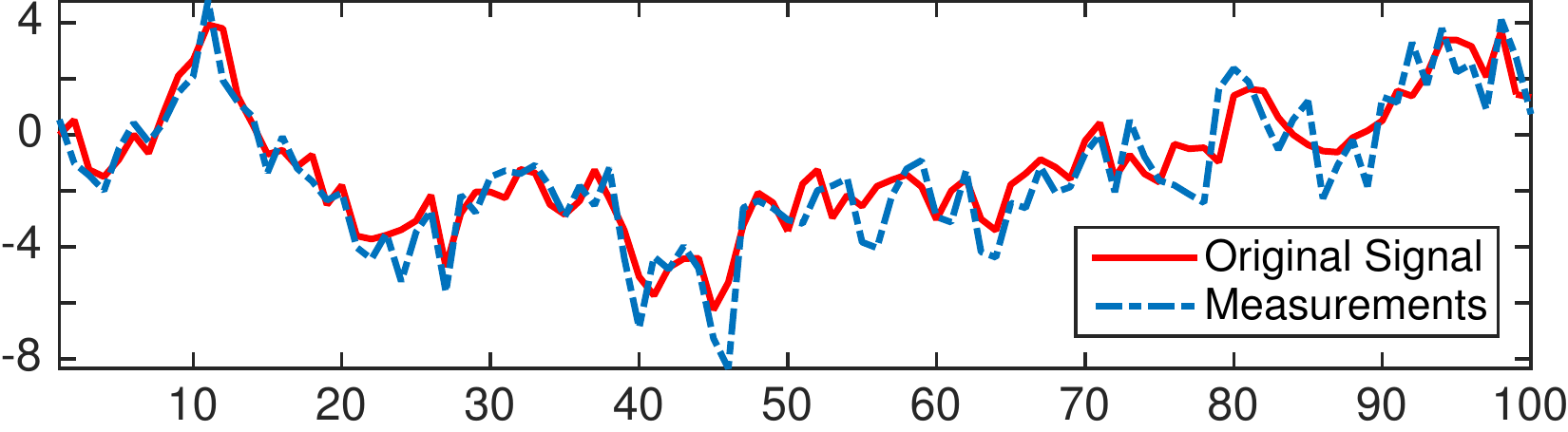}
		\caption{Brownian}
		\label{fig:realization_Brownian}
	\end{subfigure}
	\\
	\begin{subfigure}[b]{0.9\linewidth}
		\centering
		\includegraphics[width=\linewidth]{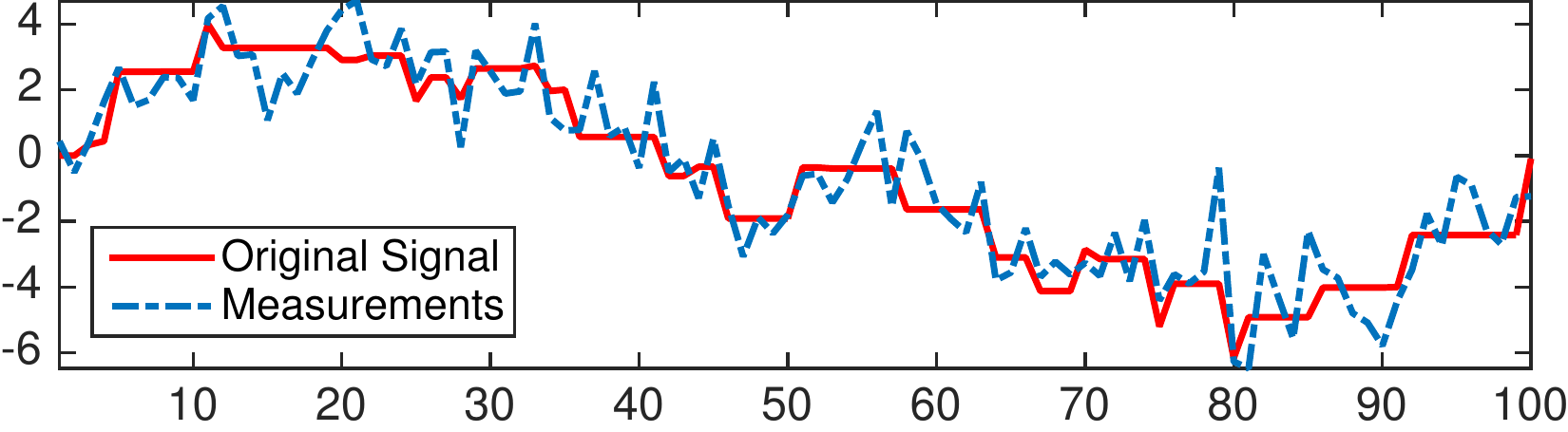}
		\caption{Compound Poisson}
		\label{fig:realization_Poisson}
	\end{subfigure}
	\caption{Realizations of a Brownian motion and a compound Poisson process are plotted along with their corrupted versions with AWGN of variance $\sigma^2=1$.}
	\label{fig:realizations}
\end{figure}

In this section, we report the experimental denoising results of the two proposed learning schemes: ADMM with \emph{unconstrained} shrinkage functions learned via Algorithm~\ref{alg:learn_uncons} (denoted MMSE-ADMM) and ADMM with \emph{constrained} shrinkage functions learned via Algorithm~\ref{alg:learn_cons} (denoted MMSE-CADMM). 
Throughout this section, the transform $\bL$ is fixed to be the finite difference operator: $[\bL\bx]_i = x_i - x_{i-1},\forall i$; the signal length is fixed to $N=100$. Experiments were implemented in MATLAB on the two following types of L\'{e}vy processes:
\begin{enumerate}
	\item Brownian motion: entries of the increment vector $\bu=\bL\bx$ are i.i.d. Gaussian with unit variance: $p_U(u) =  \Frac{e^{-\Frac{u^2}{2}}}{\sqrt{2\pi}}$.
	\item Compound Poisson: entries of the increment vector $\bu=\bL\bx$ are i.i.d. Bernoulli-Gaussian:
	$p_U(u) =  (1-e^{-\lambda})\,\Frac{e^{-\Frac{u^2}{2}}}{\sqrt{2\pi}} + e^{-\lambda}\delta(u)$, where  $\delta$ is the Dirac impulse and $\lambda=0.6$ is fixed. This results in a piecewise-constant signal $\bx$ with Gaussian jumps.
\end{enumerate}
Specific realizations of these processes and their corrupted versions with AWGN of variance $\sigma^2=1$ are shown in~Fig.~\ref{fig:realizations}.

\subsection{Denoising performance}
The same parameters were chosen for both learning schemes (constrained and unconstrained). In particular, for each type of processes, a set of 500 signals was used for training and another set of 500 signals was used for testing. The number of ADMM layers (iterations) was set to $K=10$; the penalty parameter of the augmented Lagrangian was set to $\mu=2$.
The shrinkage function was represented with the cubic B-spline:
\begin{align*}
\psi(x)=\beta^{(3)}(x) = 
\begin{cases}
\frac{2}{3} - |x|^2  + \frac{|x|^3}{2}, & 0\leq |x| < 1\\
\frac{1}{6} \left(2-|x|\right)^3, & 1\leq  |x| < 2\\
0,                                & 2\leq |x|.
\end{cases}
\end{align*}
The spline coefficients $\{c_m\}$ were located uniformly in the dynamic range of $\bu=\bL\bx$ with sampling step $\Delta = \sigma/2$, which is dependent on the noise level. Learning was performed by running either Algorithm~\ref{alg:learn_uncons} or Algorithm~\ref{alg:learn_cons} for 1000 iterations with learning rate $\gamma = 2\times 10^{-4}$. The shrinkage function was always initialized with the \emph{identity} line, which corresponds to $c^{(0)}_m = m$  for all $m$. 

\begin{figure}
	\centering
	\begin{subfigure}[b]{0.48\linewidth}
		\centering
		\includegraphics[width=\linewidth]{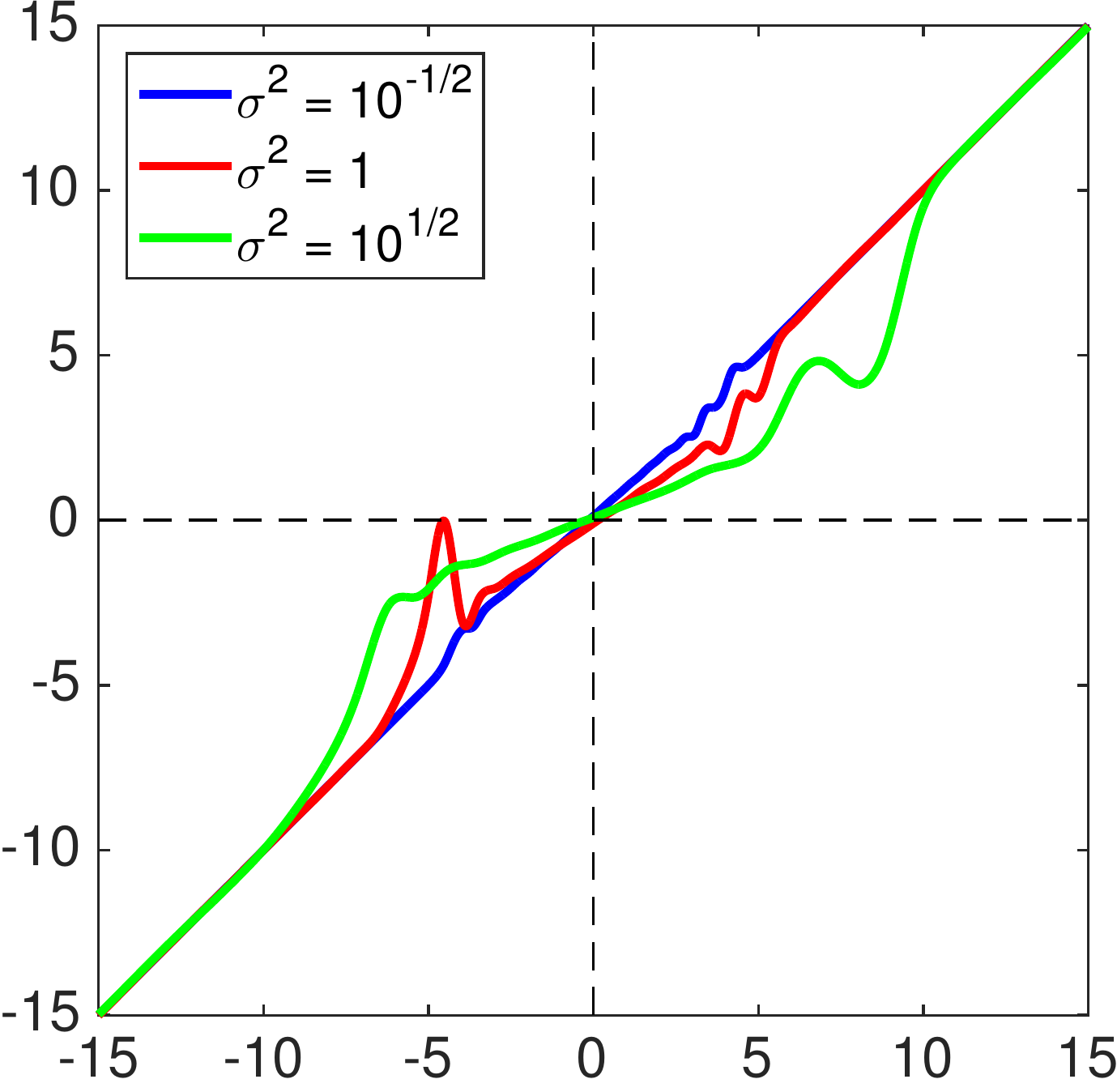}
		\caption{Brownian motion}
		\label{fig:shrink_unc_Brownian}
	\end{subfigure}
	\begin{subfigure}[b]{0.48\linewidth}
		\centering
		\includegraphics[width=\linewidth]{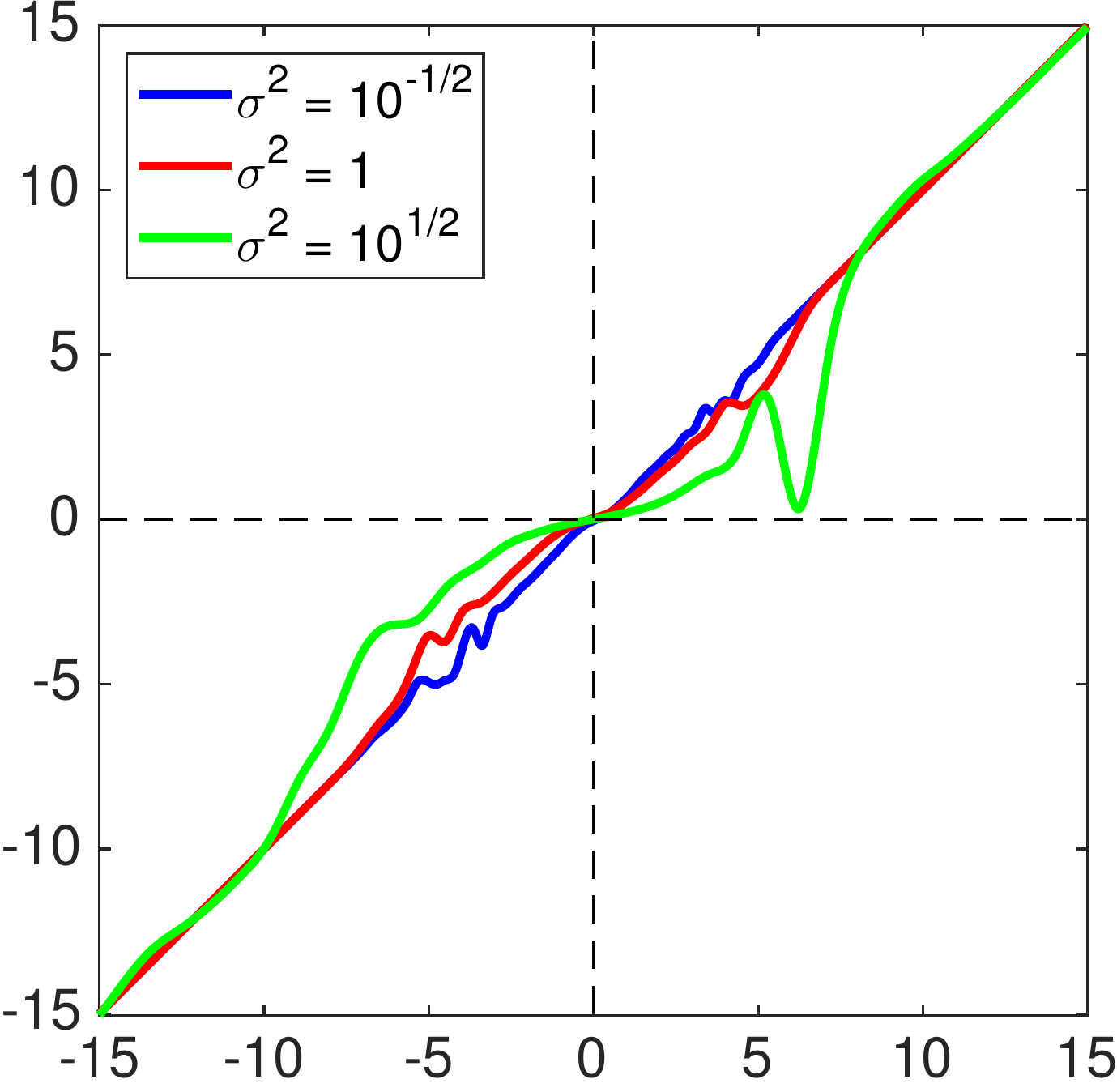}
		\caption{Compound Poisson}
		\label{fig:shrink_unc_Poisson}
	\end{subfigure}
	\caption{Unconstrained shrinkage functions learned from data for various noise variances $\sigma^2$.}
	\label{fig:shrinkages_unc}
\end{figure}

\begin{figure}
	\centering
	\begin{subfigure}[b]{0.48\linewidth}
		\centering
		\includegraphics[width=\linewidth]{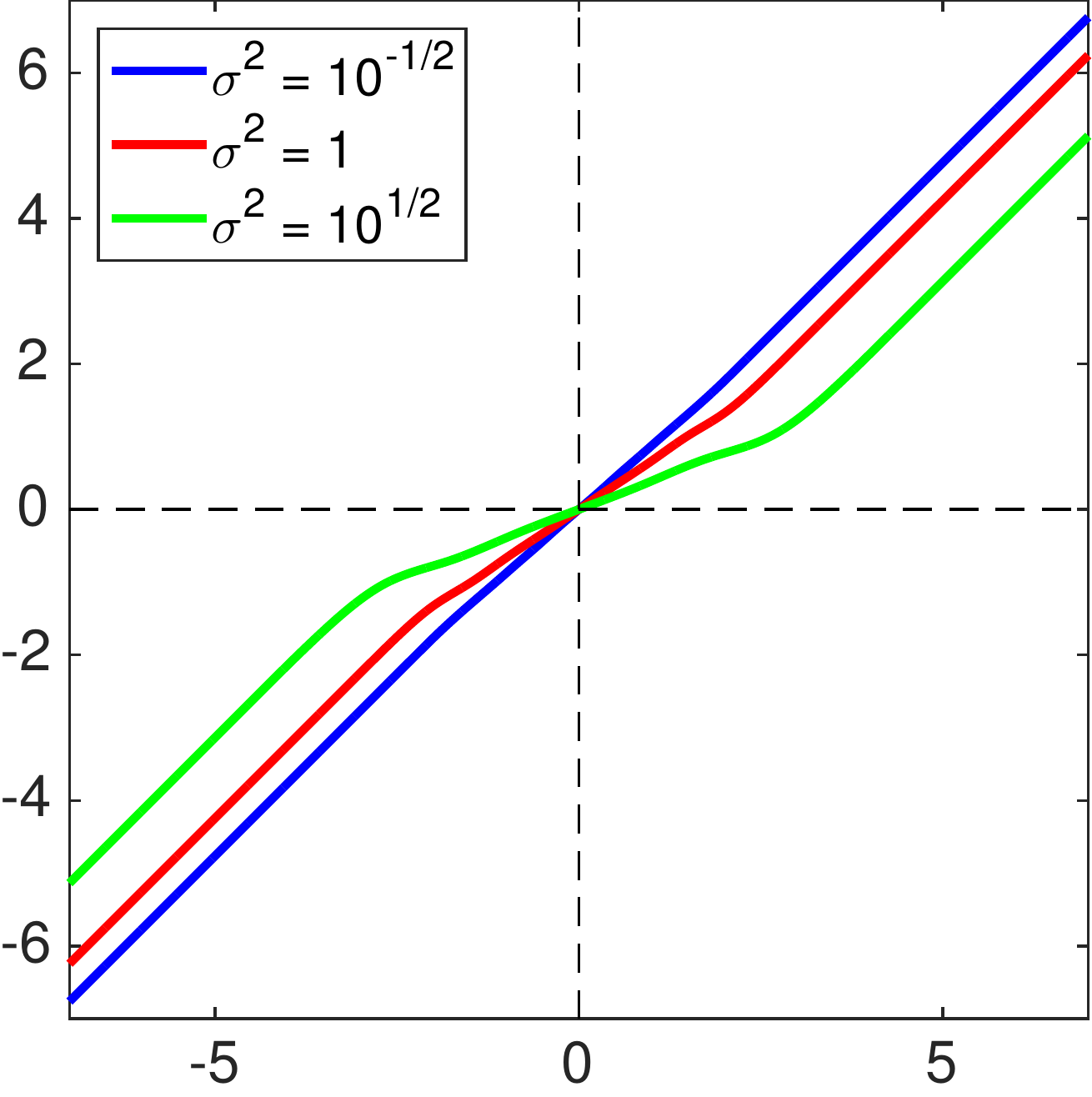}
		\caption{Brownian motion}
		\label{fig:shrink_con_Brownian}
	\end{subfigure}
	\begin{subfigure}[b]{0.48\linewidth}
		\centering
		\includegraphics[width=\linewidth]{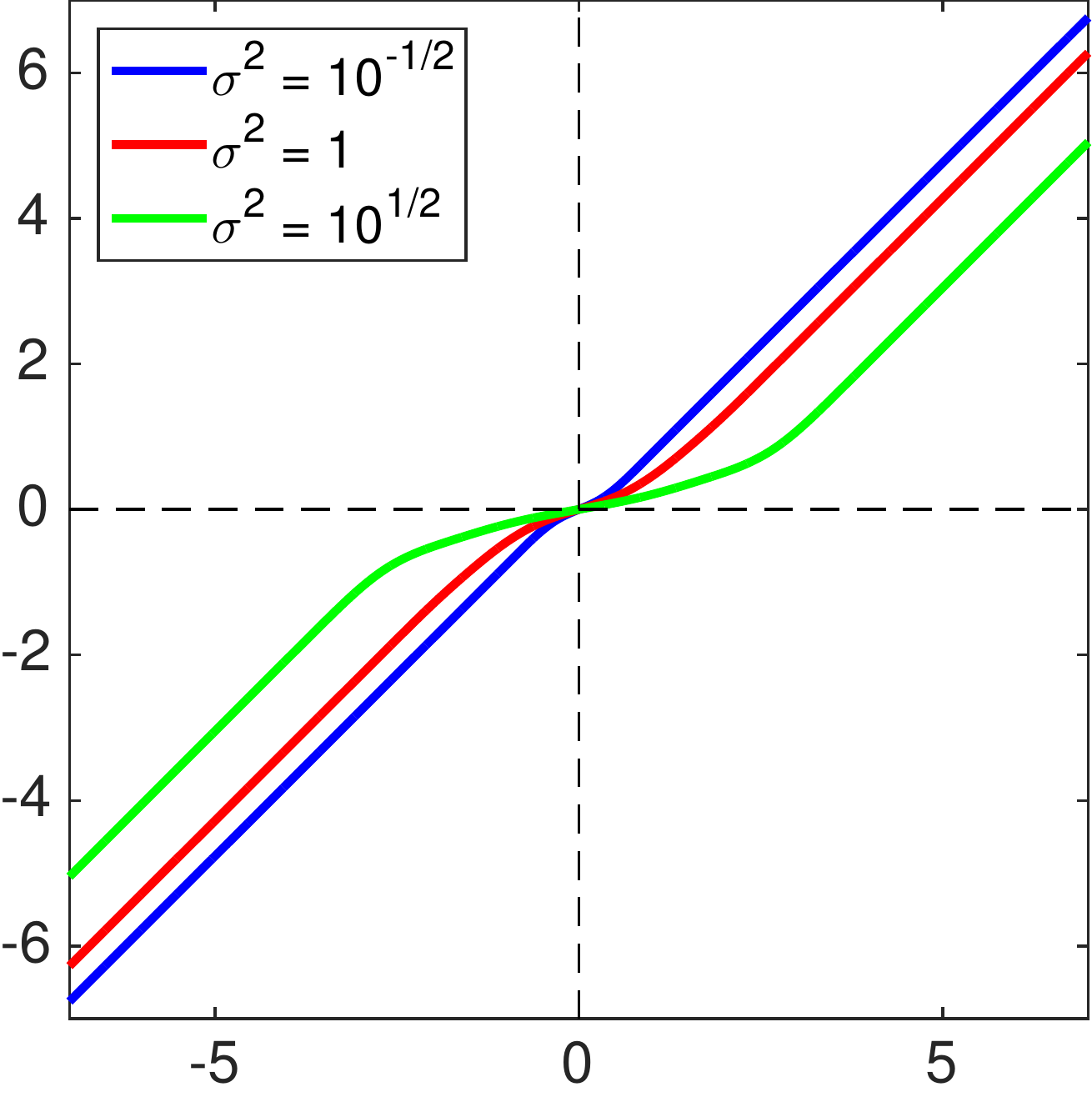}
		\caption{Compound Poisson}
		\label{fig:shrink_con_Poisson}
	\end{subfigure}
	\caption{Antisymmetric and firmly nonexpansive shrinkage functions learned from data for various noise variances $\sigma^2$.}
	\label{fig:shrinkages}
\end{figure}

\begin{figure}
	\centering
	\begin{subfigure}[b]{0.48\linewidth}
		\centering
		\includegraphics[width=\linewidth]{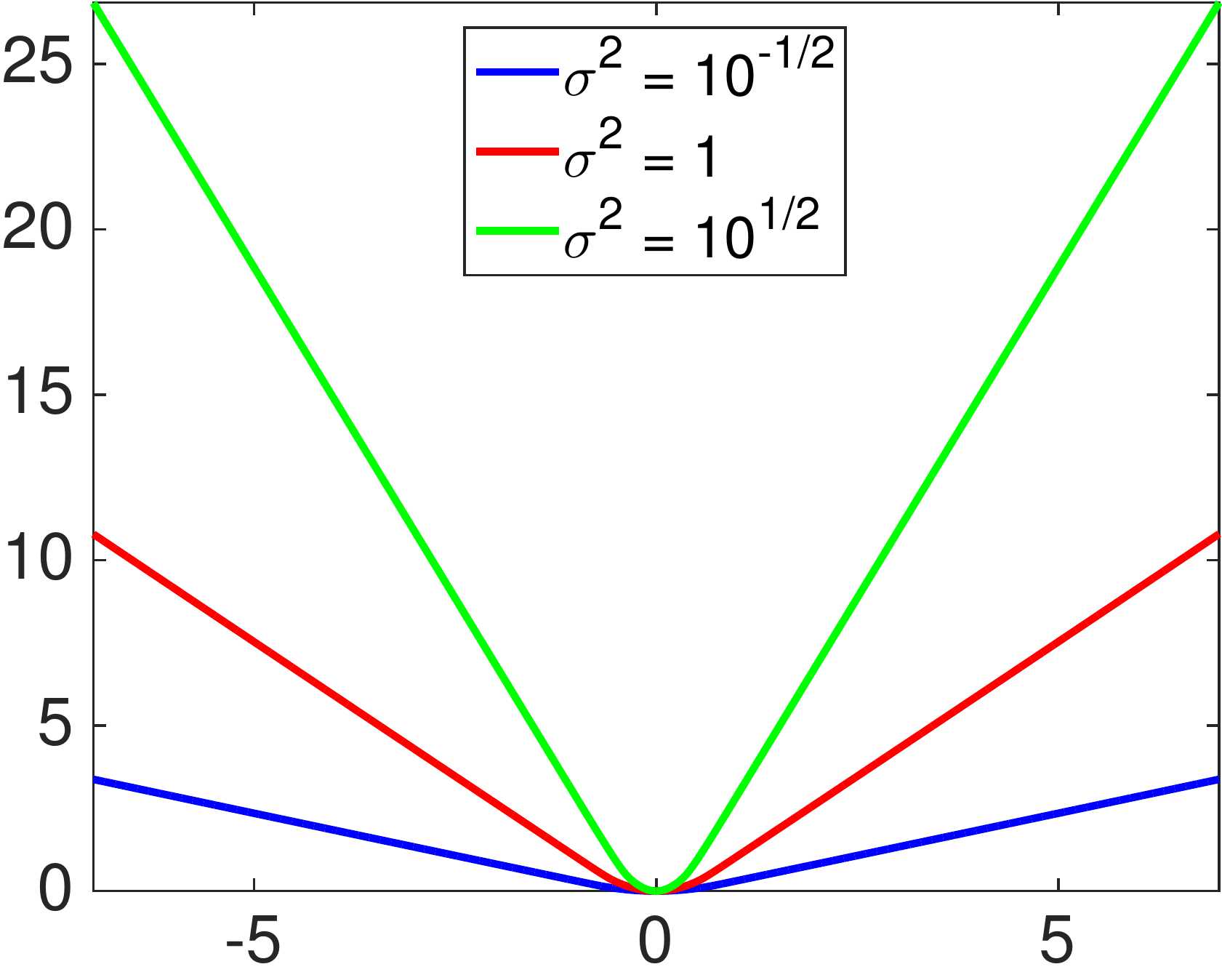}
		\caption{Brownian motion}
		\label{fig:potential_con_Brownian}
	\end{subfigure}
	\begin{subfigure}[b]{0.48\linewidth}
		\centering
		\includegraphics[width=\linewidth]{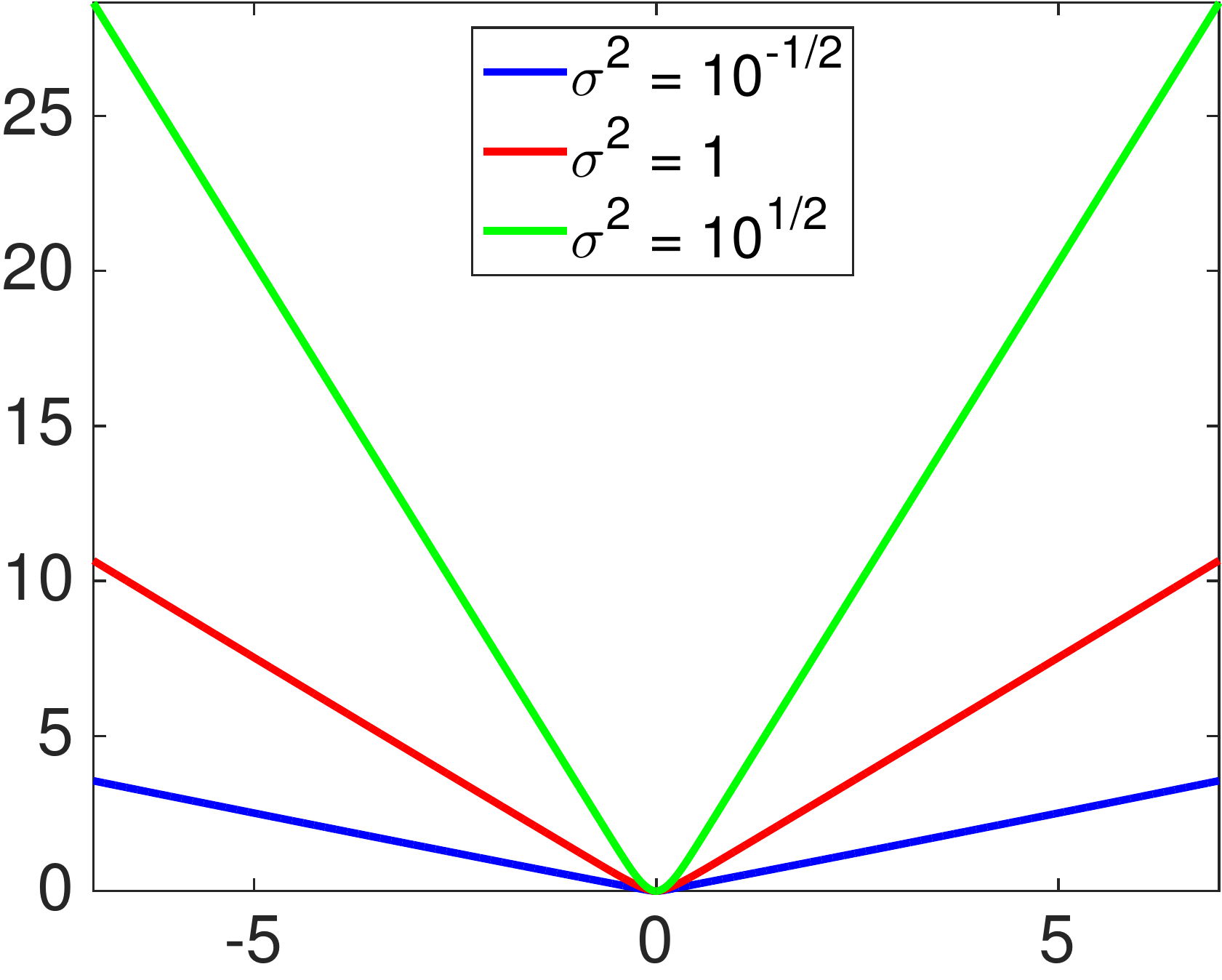}
		\caption{Compound Poisson}
		\label{fig:potential_con_Poisson}
	\end{subfigure}
	\caption{Symmetric and convex penalty functions that admit the learned constrained shrinkage functions in Fig.~\ref{fig:shrinkages} as their proximal operators for various noise variances $\sigma^2$.}
	\label{fig:potentials}
\end{figure}

The denoising performances were numerically evaluated by the signal-to-noise ratio (SNR) improvement
that is defined by
$
\Delta {\rm SNR}\,[\rm dB] =  10 \log_{10} \left(\Frac{\|\hat{\bx}-\bx\|^2_2}{\|\by-\bx\|^2_2}\right).
$
We compare the results of MMSE-ADMM and MMSE-CADMM against the following reconstruction methods:
\begin{enumerate}
	\item MMSE: This is the optimal estimator (in the MSE sense) and is obtained through  a message-passing algorithm~\cite{KamilovPAU:2013}.
	\item LMMSE (Linear MMSE): The best linear estimation is obtained by applying the Wiener filter to the noisy observation: $\hat{\bx}_{\rm LMMSE}=(\bI + \sigma^2\bL^T\bL)^{-1}\by$. This is also the least-square solution with $\ell_2$ (Tikhonov-like) regularization.  
	\item TV (Total Variation)~\cite{RudinOF:1992}: This estimator is obtained with an $\ell_1$ regularizer whose proximal operator is simply a soft-thresholding: $T_{\lambda}(u)=1_{\{|u|>\lambda\}} {\rm sign}(u)(|u|-\lambda)$. In our experiments, the regularization parameter $\lambda$ is \emph{optimized} for each signal.
\end{enumerate}
\begin{figure}
	\centering
	\begin{subfigure}[b]{0.9\linewidth}
		\centering
		\includegraphics[width=\linewidth]{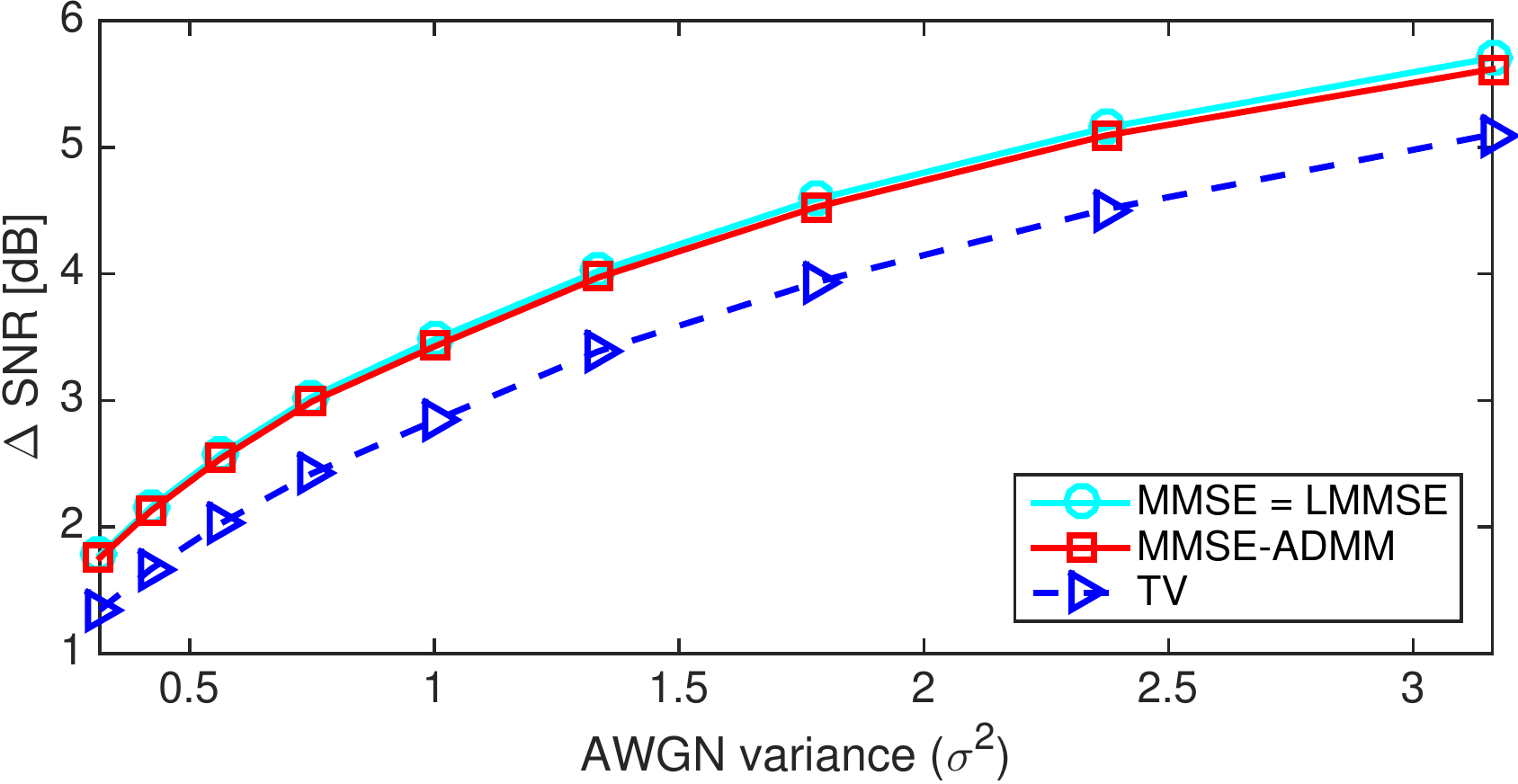}
		\caption{Brownian}
		\label{fig:Brownian_unconstrained}
	\end{subfigure}
	\\
	\begin{subfigure}[b]{0.9\linewidth}
		\centering
		\includegraphics[width=\linewidth]{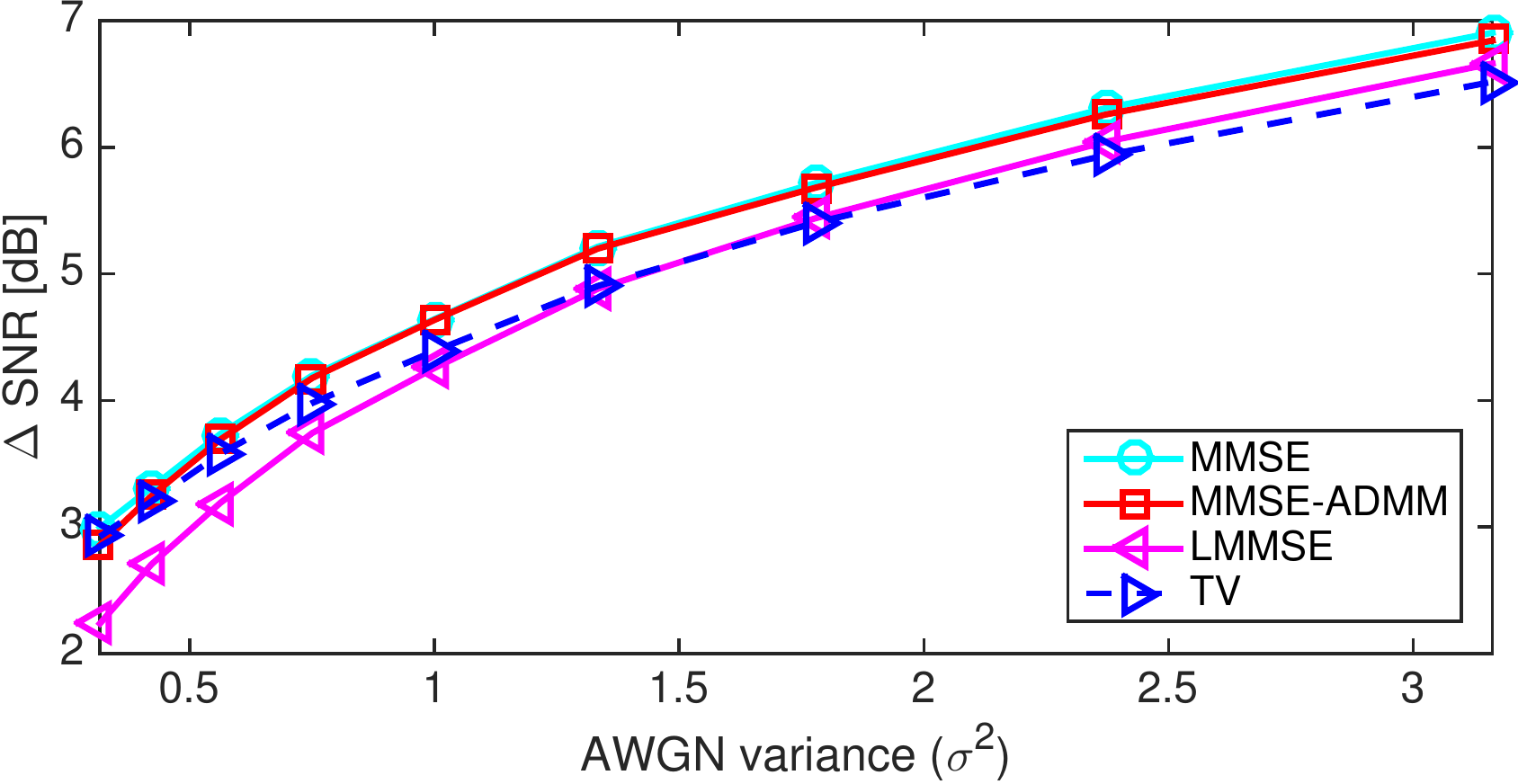}
		\caption{Compound Poisson}
		\label{fig:CompoundPoisson_unconstrained}
	\end{subfigure}
	\caption{Denoising performances of the MMSE-ADMM where the unconstrained shrinkage functions are learned for all instances of the noise variance.}
	\label{fig:learning_unconstrained}
\end{figure}
\begin{figure}
	\centering
	\begin{subfigure}[b]{0.9\linewidth}
		\centering
		\includegraphics[width=\linewidth]{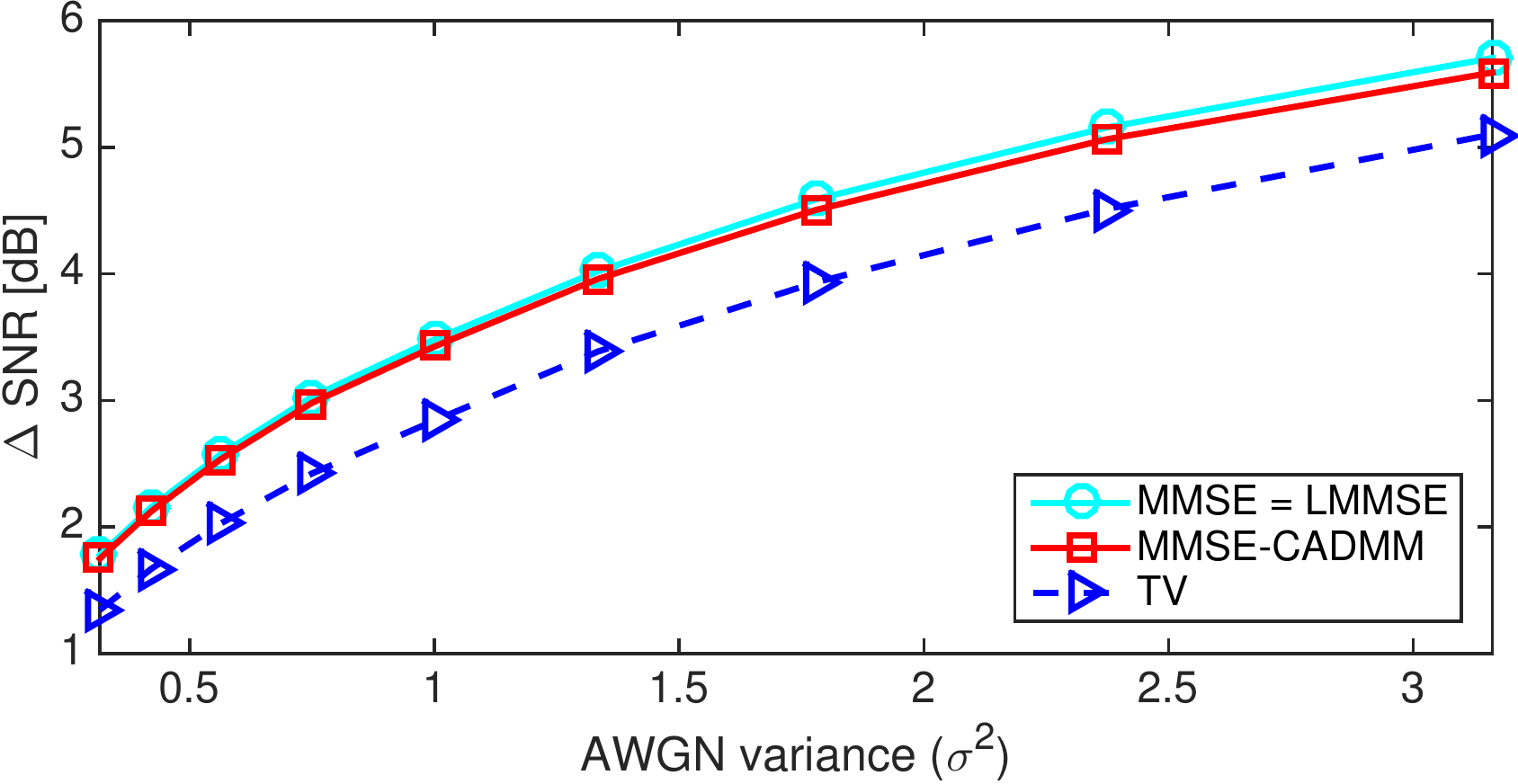}
		\caption{Brownian}
		\label{fig:Brownian}
	\end{subfigure}
	\\
	\begin{subfigure}[b]{0.9\linewidth}
		\centering
		\includegraphics[width=\linewidth]{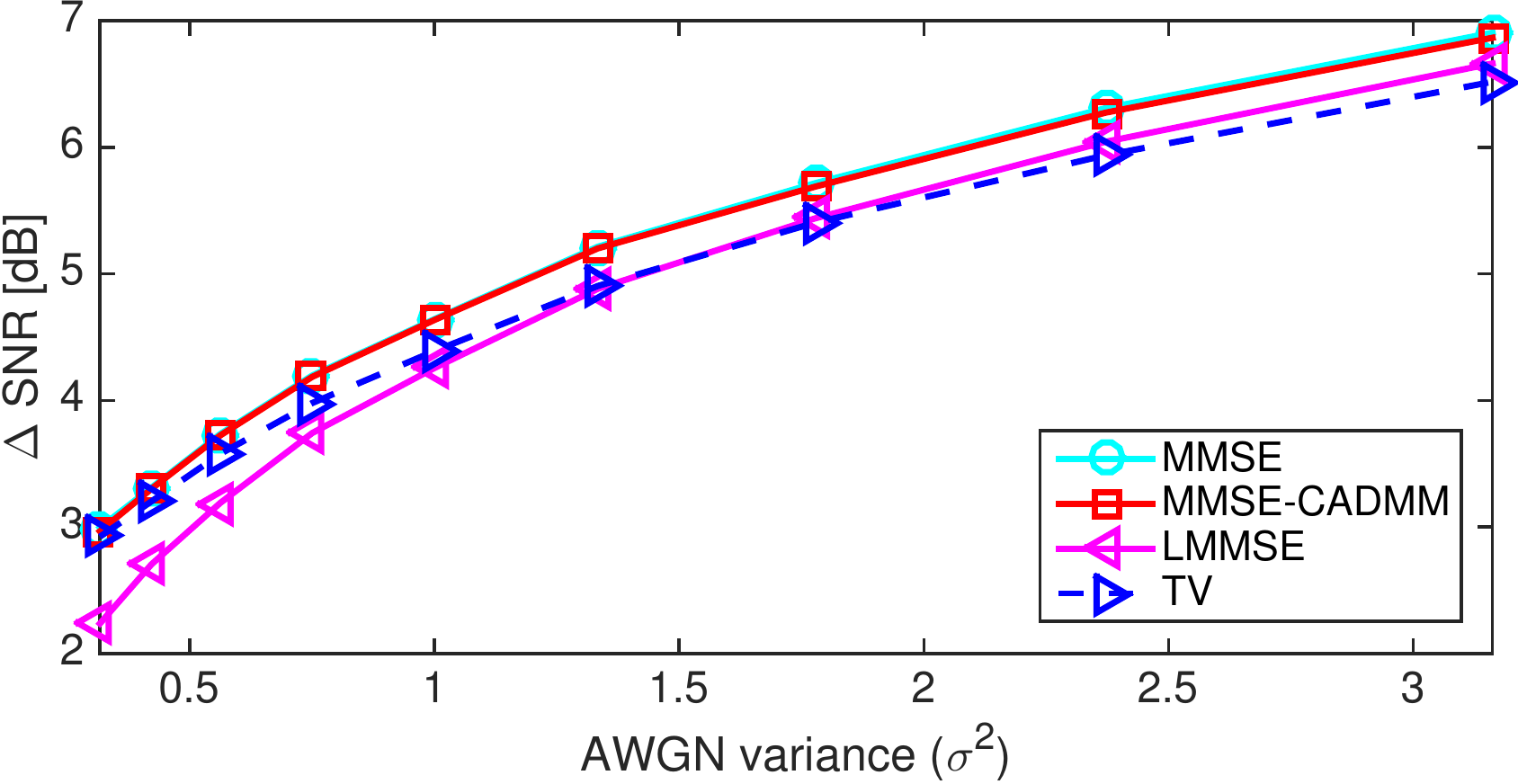}
		\caption{Compound Poisson}
		\label{fig:CompoundPoisson}
	\end{subfigure}
	\caption{Denoising performances of the MMSE-CADMM where the constrained shrinkage functions are learned for all instances of the noise variance.}
	\label{fig:learning}
\end{figure}
The denoising performances of MMSE-ADMM and MMSE-CADMM for various noise variances between $10^{-\Frac{1}{2}}$ and $10^{\Frac{1}{2}}$ are reported in Figs.~\ref{fig:learning_unconstrained} and~\ref{fig:learning}, respectively. It is remarkable that both MMSE-ADMM and MMSE-CADMM curves are almost identical to the optimal MMSE curve (with the largest gap being about 0.1 dB) and significantly outperform TV, for both types of signals, and LMMSE, for compound Poisson processes. Note that, for Brownian motions, LMMSE and MMSE are the same. The unconstrained and constrained shrinkage functions learned for three different levels of noise are illustrated in Figs.~\ref{fig:shrinkages_unc} and~\ref{fig:shrinkages}, respectively. As can be seen in Fig.~\ref{fig:shrinkages_unc}, the unconstrained learning might result in non-monotonic curves, which cannot be the proximal operators of any penalty functions, according to~\cite[Proposition 1]{SchmidtR:2014}. By contrast, the antisymmetric and firmly nonexpansive curves in Fig.~\ref{fig:shrinkages} are the proximal operators of the symmetric and convex penalty functions that are plotted in Fig.~\ref{fig:potentials}. These functions were numerically obtained by integrating $\partial\Phi=(T-\id)^{-1}$, where $T$ is the learned shrinkage function.

\subsection{Constrained versus unconstrained learning}
\begin{figure}[h]
	\centering
	\begin{subfigure}[b]{0.9\linewidth}
		\centering
		\includegraphics[width=\linewidth]{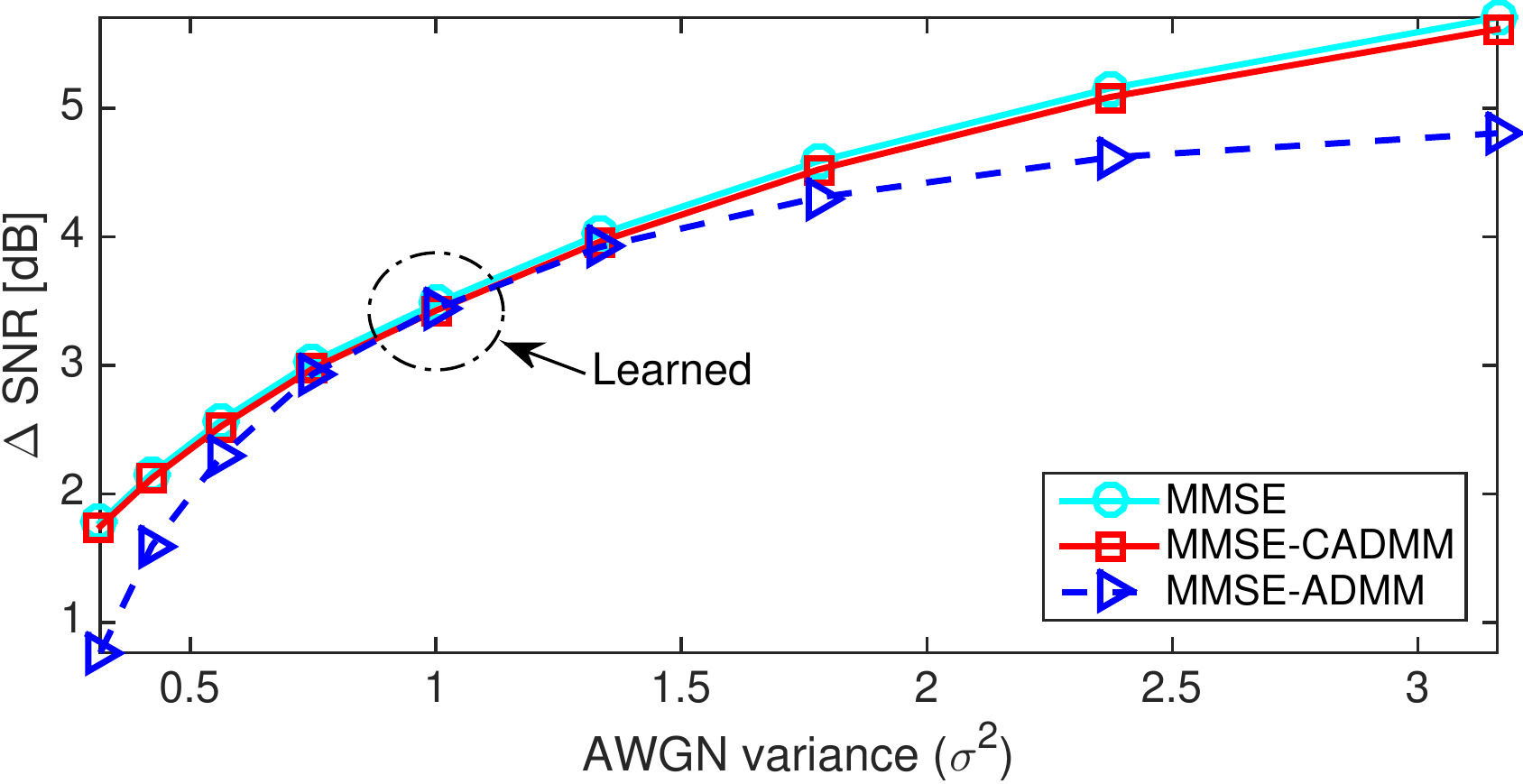}
		\caption{Brownian}
		\label{fig:SNR_Brownian_learn_once}
	\end{subfigure}
	\\
	\begin{subfigure}[b]{0.9\linewidth}
		\centering
		\includegraphics[width=\linewidth]{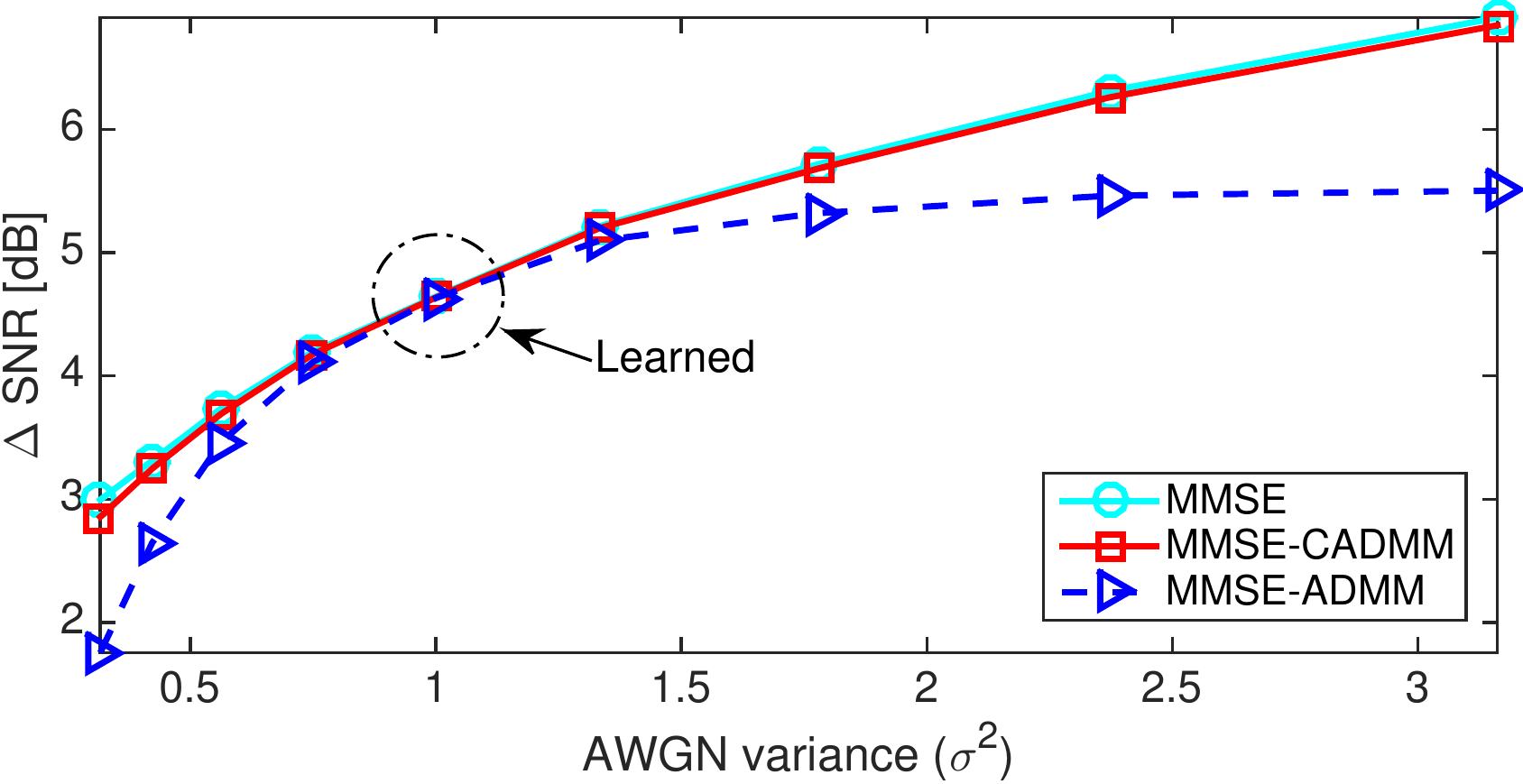}
		\caption{Compound Poisson}
		\label{fig:SNR_Poisson_learn_once}
	\end{subfigure}
	\caption{Learning once and for all: only the shrinkage function for $\sigma^2=1$ is learned (with and without constraints) and the rest are obtained by scaling the learned penalty function with corresponding values of $\sigma^2$.}
	\label{fig:learn_once}
\end{figure}

To demonstrate the benefits of the constrained learning over the unconstrained one, we compare their denoising performances for 9 different levels of noise as before, but this time only the shrinkage function $T$ for $\sigma^2=1$ was learned. For constrained learning, the shrinkage function with respect to another noise variance $\sigma^2$ was numerically computed by using the formula
\begin{align*}
T_{\sigma^2} = \left(\sigma^2T^{-1} + (1-\sigma^2)\id\right)^{-1}.
\end{align*}
For unconstrained learning, these computations are prohibited, and so the learned shrinkage function for $\sigma^2=1$ was used for all the other noise levels. The results were illustrated in Fig.~\ref{fig:learn_once}. It is noticeable that MMSE-CADMM is much better than MMSE-ADMM and, surprisingly, almost as good as the optimal MMSE for all levels of noise, even though the (constrained) learning was performed only once. In other words, the experiments suggest that the proposed MMSE-CADMM combines desired properties of the MAP and MMSE estimators: fast implementation and scalability with noise variance of MAP and optimality of MMSE.

\begin{figure}[h]
	\centering
	\includegraphics[width=0.9\linewidth]{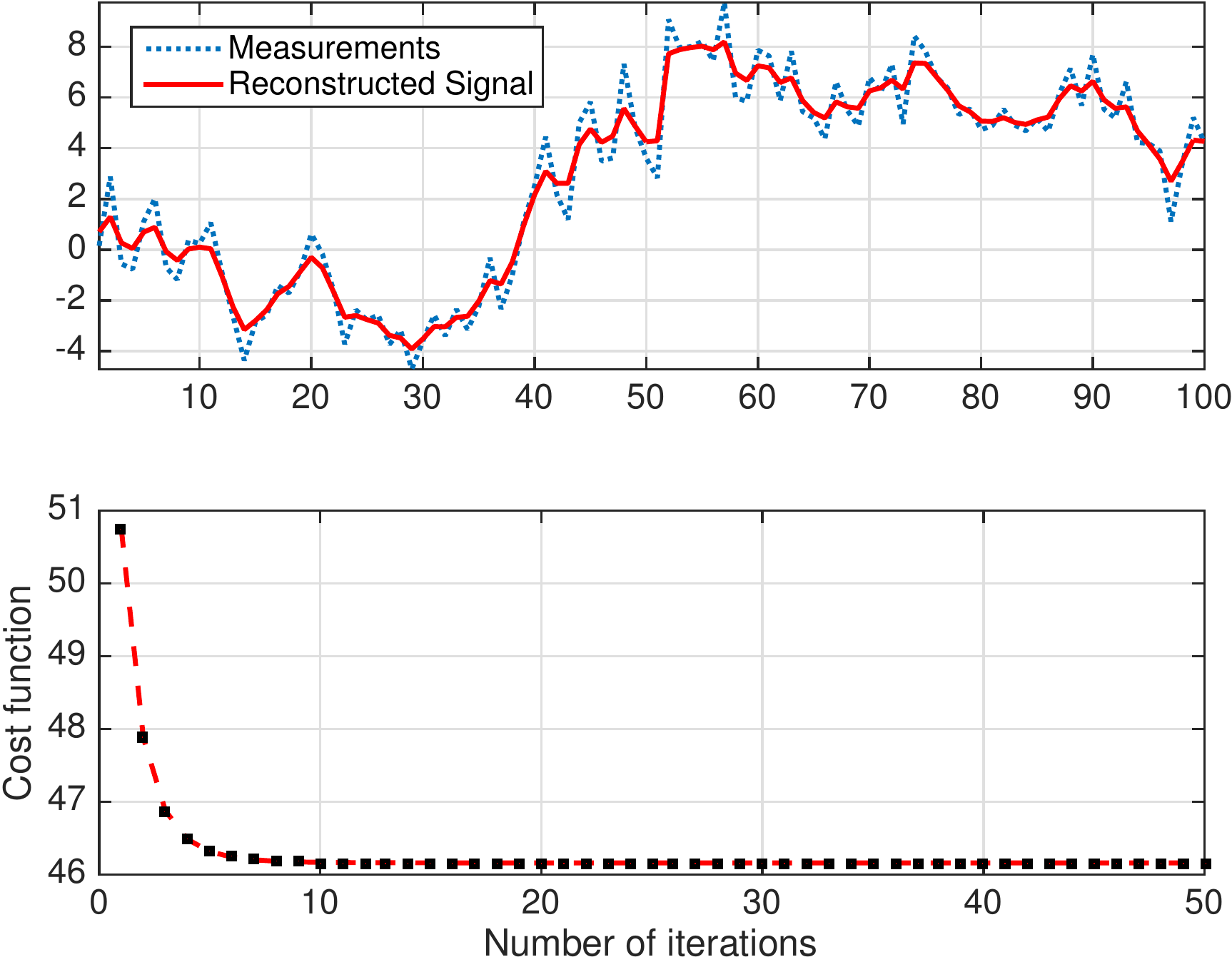}
	\caption{Reconstruction of a specific Brownian motion with AWGN of variance $\sigma^2 =1$ using MMSE-CADMM. Values of the underlying cost function are plotted for the first 50 iterations of ADMM.}
	\label{fig:cost_Brownian}
\end{figure}

\begin{figure}[h]
	\centering
	\includegraphics[width=0.9\linewidth]{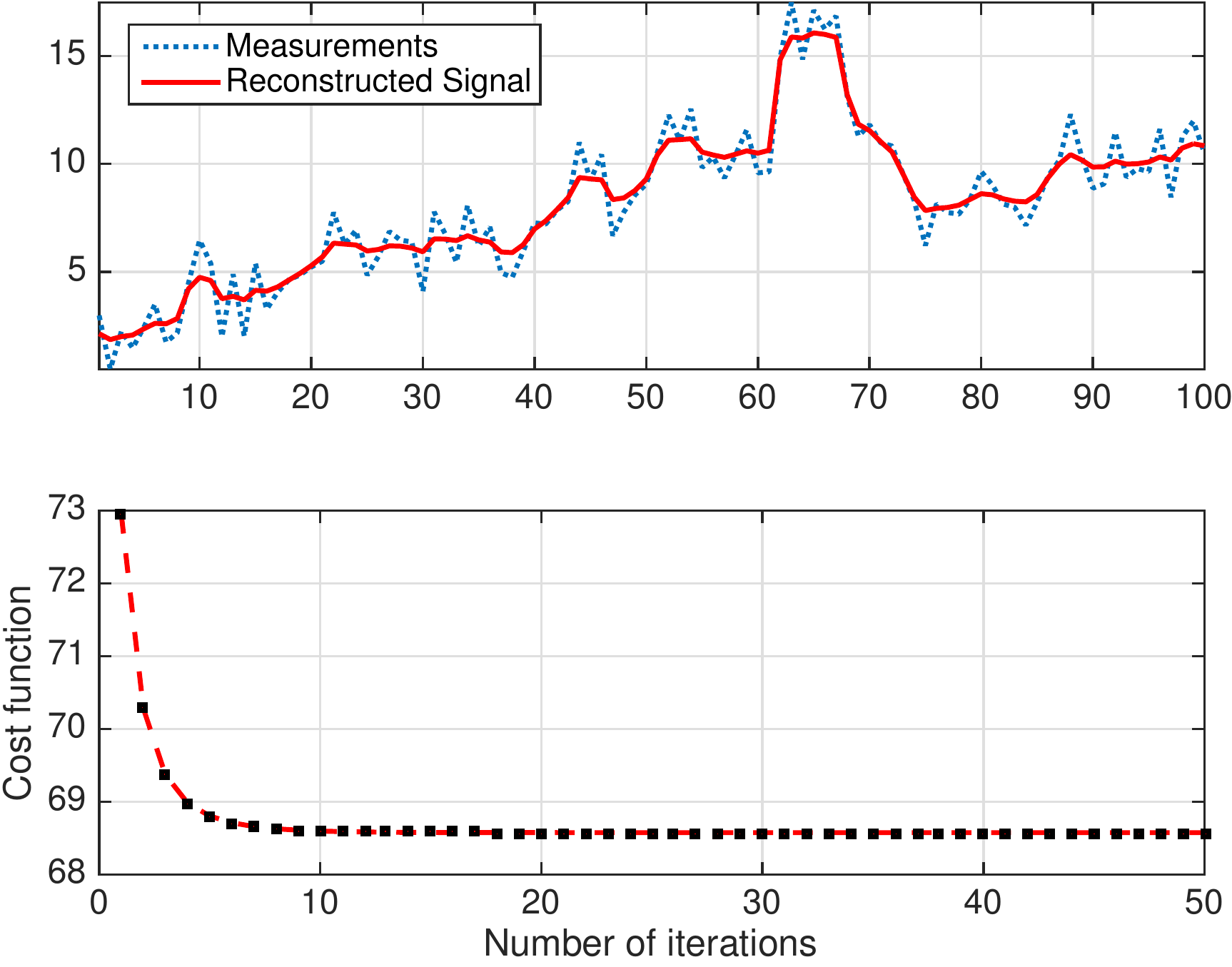}
	\caption{Reconstruction of a specific compound Poisson signal with AWGN of variance $\sigma^2 =1$ using MMSE-CADMM. Values of the underlying cost function are plotted for the first 50 iterations of ADMM.}
	\label{fig:cost_CP}
\end{figure}

\begin{figure}[h]
	\centering
	\includegraphics[width=0.9\linewidth]{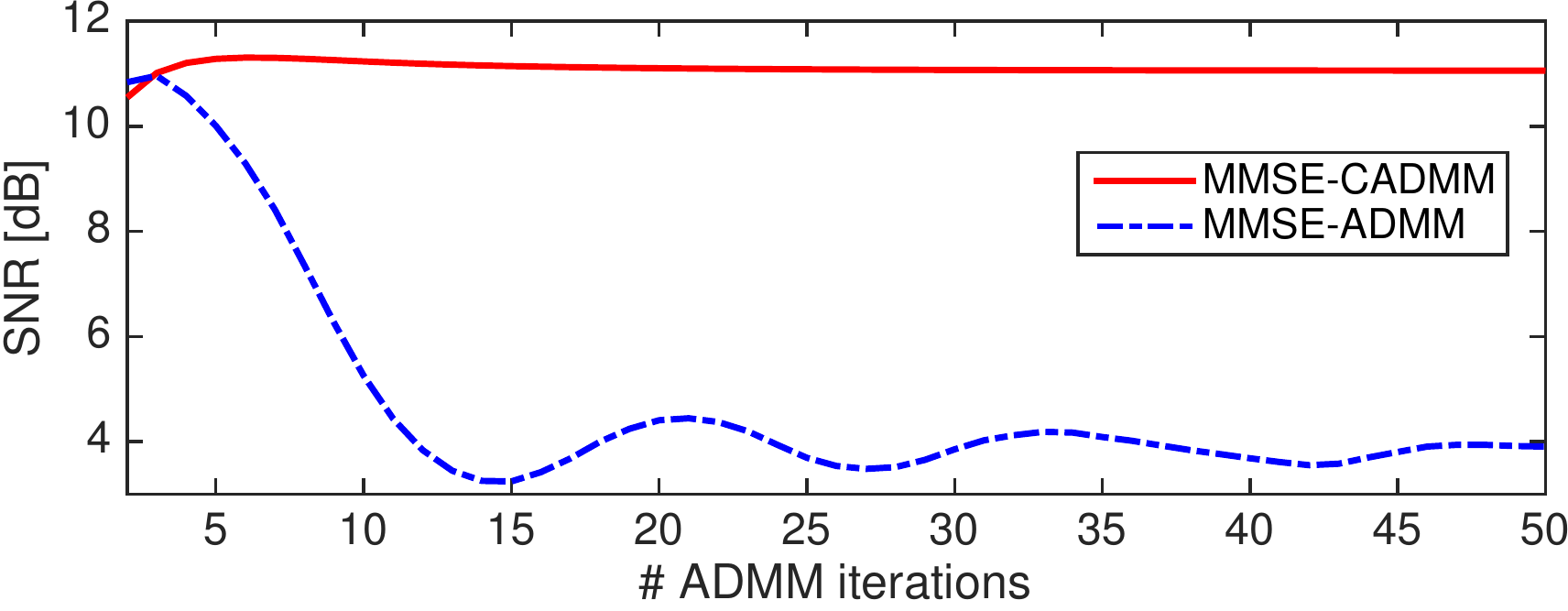}
	\caption{Average SNRs when denoising compound Poisson signals with constrained and unconstrained learning schemes are plotted against the number of ADMM iterations used in the testing phase ($K_{\rm test}$). The constrained and unconstrained shrinkage functions were both trained  with $K_{\rm train}=2$ and $\sigma^2=10$.} 
	\label{fig:con_vs_unc}
\end{figure}

Another advantage of the constrained learning is its convergence guarantee that is associated with the minimization of an underlying cost function (as mentioned in Theorem~\ref{thm:converge}), which does not necessarily exist in the case of unconstrained learning. Figs.~\ref{fig:cost_Brownian} and~\ref{fig:cost_CP} illustrates the reconstructions of a Brownian motion and a compound Poisson signal, respectively, from their noisy measurements using MMSE-CADMM, and the convergences of the corresponding cost functions. Experiments also show that the constrained learning is much more stable to the number of ADMM iterations  used in the testing phase ($K_{\rm test}$) when it is different from the number of ADMM iterations  used in the training phase ($K_{\rm train}$). Fig.~\ref{fig:con_vs_unc} demonstrates this observation by plotting the average SNRs of denoising compound Poisson signals using MMSE-ADMM and MMSE-CADMM against $K_{\rm test}$ ranging from 2 to 50. In this experiment, both constrained and unconstrained learnings were  performed with $K_{\rm train}=2$ and $\sigma^2=10$. It can be seen from the plot that, when $K_{\rm test}$ increases, the SNR of MMSE-ADMM tends to decrease and fluctuate significantly, while the SNR of MMSE-CADMM tends to improve and converge. 

\section{Conclusion}
We have developed in this paper a learning scheme for signal denoising using ADMM in which a single (iteration-independent) shrinkage function is  constrained to be antisymmetric firmly-nonexpansive and learned from data via a simple projected gradient descent to minimize the reconstruction error. This constrained shrinkage function is proved to be the proximal operator of a symmetric convex penalty function. Imposing constraints on the shrinkage function gains several striking advantages: the antisymmetry reduces the number of learning parameters by a half, while the firm nonexpansiveness guarantees the convergence of ADMM, as well as the scalability with noise level. Yet, the denoising performance of the proposed learning scheme is empirically identical to the optimal MMSE estimators for the two types of L\'{e}vy processes in a wide range of noise variances. Our experiments also demonstrate that learning the convex penalty function for one level of noise (via learning its proximal operator) and then scaling it for other noise levels yields equivalent performances to those of direct leaning for all noise levels. This property opens up an opportunity to vastly improve the robustness and generalization ability of learning schemes. Potential directions for future research include extension of the proposed framework to general inverse problems of the form $\by = \bH\bx+\bn$ as well as to multidimensional signals. Another issue worth investigating is the joint learning of the shrinkage function and the decorrelation (sparsifying) transform $\bL$ from real data, like images, whose statistics are unknown.

\appendices
\section{Proof of Theorem~\ref{thm:symmetry}} \label{app:symmetry}
	We first recall that
	\begin{align}\label{eq:resolvent}
	\prox_\Phi =(\partial\Phi + \id)^{-1}.
	\end{align}
	Assume for now that $\Phi$ is symmetric. Fix $\bx\in\RR$ and let $\bu = \prox_{\Phi}(\bx)$, $\bv = \prox_{\Phi}(-\bx)$. We need to show that $\bu=-\bv$.
	From~\eqref{eq:resolvent}, we have that
	\begin{align*}
	\bx - \bu &\in \partial\Phi(\bu),\\
	-\bx - \bv &\in \partial\Phi(\bv).
	\end{align*}
	By the definition of the subdifferential operator, we obtain the following inequalities: 
	\begin{align}
	\Phi(-\bv) - \Phi(\bu) &\geq \ip{\bx-\bu}{-\bv-\bu}\\
	\Phi(-\bu) - \Phi(\bv) &\geq \ip{-\bx-\bv}{-\bu-\bv},
	\end{align}
	which, by the symmetry of $\Phi$, can be further simplified to
	\begin{align}
	\Phi(\bv) - \Phi(\bu) &\geq \ip{\bu-\bx}{\bu+\bv}\\
	\Phi(\bu) - \Phi(\bv) &\geq \ip{\bx+\bv}{\bu+\bv}.
	\end{align}
	Adding these inequalities yields $\norm{\bu+\bv}^2_2\leq 0$, or $\bu=-\bv$.
	
	Assume conversely that $\prox_\Phi$ is antisymmetric. We first show that $\bu\in\partial\Phi(\bx)$ is equivalent to $-\bu\in\partial\Phi(-\bx)$. Indeed, by using~\eqref{eq:resolvent} and from the antisymmetry of $\prox_\Phi$, 
	\begin{align*}
	\bu\in\partial\Phi(\bx) &\Leftrightarrow \bu + \bx \in \partial\Phi(\bx) + \bx = \prox_\Phi^{-1}(\bx)\\
	&\Leftrightarrow \bx = \prox_\Phi(\bu+\bx)\\
	&\Leftrightarrow -\bx = \prox_\Phi(-\bu-\bx)\\
	&\Leftrightarrow -\bu-\bx \in \prox^{-1}_\Phi(-\bx)=\partial\Phi(-\bx)-\bx\\
	&\Leftrightarrow -\bu \in \partial\Phi(-\bx).
	\end{align*}
	Furthermore, $\prox_{\Phi}(\zv)=\zv$ due to the antisymmetry. Since $\partial\Phi(\zv)=\prox^{-1}_{\Phi}(\zv)$, it must be that $\zv\in\partial\Phi(\zv)$. Let $G= \gr (\partial\Phi)$ and choose $(\bx_0,\bu_0)=(\zv,\zv)\in G$. Consider the Rockafellar anti-derivative~\cite{Rockaffellar:1997} of $\partial \Phi$:
	\begin{align}
	f(\bx) &= \sup_{n\geq 1} \sup_{\substack{(\bx_1,\bu_1)\in G\\ \cdots\\ (\bx_n,\bu_n)\in G}} \{\ipn{\bx-\bx_n}{\bu_n}+\sum_{i=0}^{n-1}\ipn{\bx_{i+1}-\bx_{i}}{\bu_i}\}\nonumber\\
	&= \sup_{n\geq 1} \sup_{\substack{(\bx_1,\bu_1)\in G\\ \cdots\\ (\bx_n,\bu_n)\in G}} \{\ipn{\bx-\bx_n}{\bu_n}+\sum_{i=1}^{n-1}\ipn{\bx_{i+1}-\bx_{i}}{\bu_i}\}\label{eq:anti-deriv}
	\end{align}
	It is well known~\cite[Proposition 22.15]{BauschkeC:2011} that $f\in\Gamma_0(\RR)$ and $\partial f = \partial \Phi$. Therefore, we can invoke~\cite[Proposition 22.15]{BauschkeC:2011} to deduce that $\Phi = f + c$, for some constant $c\in\R$. To show the symmetry of $\Phi$, it suffices to show the symmetry of $f$. From~\eqref{eq:anti-deriv} and by the symmetry of $G$, $f(-\bx)$ is equal to
	\begin{align*}
	&\sup_{n\geq 1} \sup_{\substack{(\bx_1,\bu_1)\in G\\ \cdots\\ (\bx_n,\bu_n)\in G}} \ipn{-\bx-\bx_n}{\bu_n}+\sum_{i=1}^{n-1}\ipn{\bx_{i+1}-\bx_{i}}{\bu_i}\\
	&	= \sup_{n\geq 1} \sup_{\substack{(\bx_1,\bu_1)\in G\\ \cdots\\ (\bx_n,\bu_n)\in G}} \ipn{-\bx+\bx_n}{-\bu_n}+\sum_{i=1}^{n-1}\ipn{-\bx_{i+1}+\bx_{i}}{-\bu_i}\\
	&= \sup_{n\geq 1} \sup_{\substack{(\bx_1,\bu_1)\in G\\ \cdots\\ (\bx_n,\bu_n)\in G}} \ipn{\bx-\bx_n}{\bu_n}+\sum_{i=1}^{n-1}\ipn{\bx_{i+1}-\bx_{i}}{\bu_i}\\
	&= f(\bx), \quad \forall \bx\in\RR,
	\end{align*}
	which shows that $f$ is symmetric, completing the proof.

\bibliographystyle{IEEEtran}
\bibliography{levy_denoise_lib}

\end{document}